%% file: iclr2026/iclr2026_conference.tex
\documentclass{article} 
\usepackage{iclr2026/iclr2026_conference}
\usepackage{times}

\input{iclr2026/math_commands}

\usepackage{amsfonts}       
\usepackage{amsmath}
\usepackage{amssymb}
\usepackage{amsthm}
\usepackage{array}
\usepackage{algorithm}
\usepackage{algpseudocode}
\usepackage{booktabs}       
\usepackage{bm}
\usepackage{colortbl}
\usepackage{caption}
\usepackage{float}
\usepackage{hyperref}       
\usepackage[capitalize,noabbrev]{cleveref}
\usepackage[utf8]{inputenc}
\usepackage[T1]{fontenc}    
\usepackage{graphics}       
\usepackage[utf8]{inputenc} 
\usepackage{microtype}      
\usepackage{mathtools}
\usepackage{multirow}
\usepackage{nicefrac}       
\usepackage{subfigure}
\usepackage{subcaption} 
\usepackage{tabularx}
\usepackage{url}            
\usepackage{wrapfig}
\usepackage{pifont}
\usepackage[utf8]{inputenc}  
\usepackage{textcomp} 
\usepackage{marvosym}
\usepackage{tikz}
\usetikzlibrary{arrows.meta, positioning, calc}

\title{KBVQ-MoE: KLT-guided SVD with Bias-Corrected Vector Quantization for MoE Large Language Models}

\author{
Zukang Xu$^{1}$\thanks{Equal contribution} \hfill
Zhixiong Zhao$^{1,2}$\footnotemark[1]\hspace{1ex}\thanks{This work was conducted during his internship at Houmo AI} \hfill
Xing Hu$^{1}$ \hfill
Zhixuan Chen$^{1}$ \hfill
Dawei Yang$^{1}$\thanks{Corresponding author} \\[2.0em]
\makebox[\textwidth][c]{
$^{1}$ Houmo AI \hspace{3em}
$^{2}$ Nanyang Technological University
}
}

\iclrfinalcopy

\makeatletter
\AtBeginDocument{%
  \pagestyle{fancy}%
  \fancyhf{}%
  \setlength{\headheight}{14pt}%
  \ificlrfinal
    \fancyhead[L]{Published as a conference paper at ICLR 2026}%
  \else
    \fancyhead[L]{Under review as a conference paper at ICLR 2026}%
  \fi
}
\makeatother
\begin{document}
\input{sec/0.paper}

\end{document}

%% file: iclr2026/math_commands.tex
\usepackage{amsmath,amsfonts,bm}

\def\eqref#1{equation~\ref{#1}}

\def\1{\bm{1}}

\DeclareMathAlphabet{\mathsfit}{\encodingdefault}{\sfdefault}{m}{sl}
\SetMathAlphabet{\mathsfit}{bold}{\encodingdefault}{\sfdefault}{bx}{n}



%% file: sec/0.paper.tex
\maketitle
\input{sec/1.abstract}
\input{sec/2.introduction}
\input{sec/3.related_work}
\input{sec/4.preliminaries}
\input{sec/5.method}

\input{sec/6.experiments}
\input{sec/7.conclusion}
\bibliography{iclr2026/iclr2026_conference.bib}
\bibliographystyle{iclr2026/iclr2026_conference}

\input{sec/9.appendix}

%% file: sec/1.abstract.tex
\begin{abstract}
Mixture of Experts (MoE) models have achieved great success by significantly improving performance while maintaining computational efficiency through sparse expert activation. However, their enormous parameter sizes and memory demands pose significant challenges for deployment in resource-constrained environments.
Vector Quantization (VQ) offers a promising approach for ultra-low-bit compression in Large Language Models (LLMs) by constructing and leveraging a codebook—where weight vectors are mapped to the most similar discrete codewords within the codebook. 
However, its direct application to MoEs suffers from significant performance degradation caused by two critical obstacles:  (1) redundant representation among experts leads to VQ repeatedly quantizing similar representations for each expert, resulting in inefficient utilization of the limited codebook capacity; and
(2) cumulative outputs bias is amplified by experts aggregation in MoE layers, leading to distributional shifts in the quantized outputs.
To this end, we propose KBVQ-MoE, a novel VQ framework to enhance extremely low-bit quantization for MoE-based LLMs. 
KBVQ-MoE integrates two novel techniques: 
(1) Input-driven redundancy elimination, where a Karhunen–Loève Transform (KLT) guided singular value decomposition (SVD) extracts and shares dominant weight components across experts. 
(2) Bias-corrected output stabilization, where vector quantization is applied to expert-specific (i.e., non-redundant) representations and the quantized outputs are corrected with channel-wise affine compensation.
Experiments on various MoE LLMs demonstrate that our KBVQ-MoE preserves accuracy substantially better than existing quantization methods. For instance, 3-bit quantization of Qwen1.5-MoE-A2.7B achieves an average accuracy of 67.99, nearly identical to the FP16 baseline of 68.07, underscoring the potential of KBVQ-MoE for efficient deployment on edge devices and other resource-constrained platforms.
\end{abstract}

%% file: sec/2.introduction.tex
\section{Introduction}
\label{sec:introduction}
Mixture-of-Experts (MoE) models have recently achieved state-of-the-art performance in natural language processing (NLP) \citep{gpt4, kimivl, deepseekv2, mixtral, qwen3}. By activating only a small subset of experts through a gating mechanism, MoE enables near-linear scaling of capacity with the number of experts while keeping inference cost manageable. 
However, the growth in expert count substantially increases parameter storage and memory bandwidth requirements. 
For instance, Qwen3-Next-80B-A3B\citep{qwen3-next} requires more than 160GB of GPU memory under FP16 inference. These extreme resource requirements make deployment on edge devices largely infeasible.
\begin{figure}[h!]
    \centering
    \begin{minipage}{0.38\columnwidth}
        \centering
        \includegraphics[width=\linewidth]{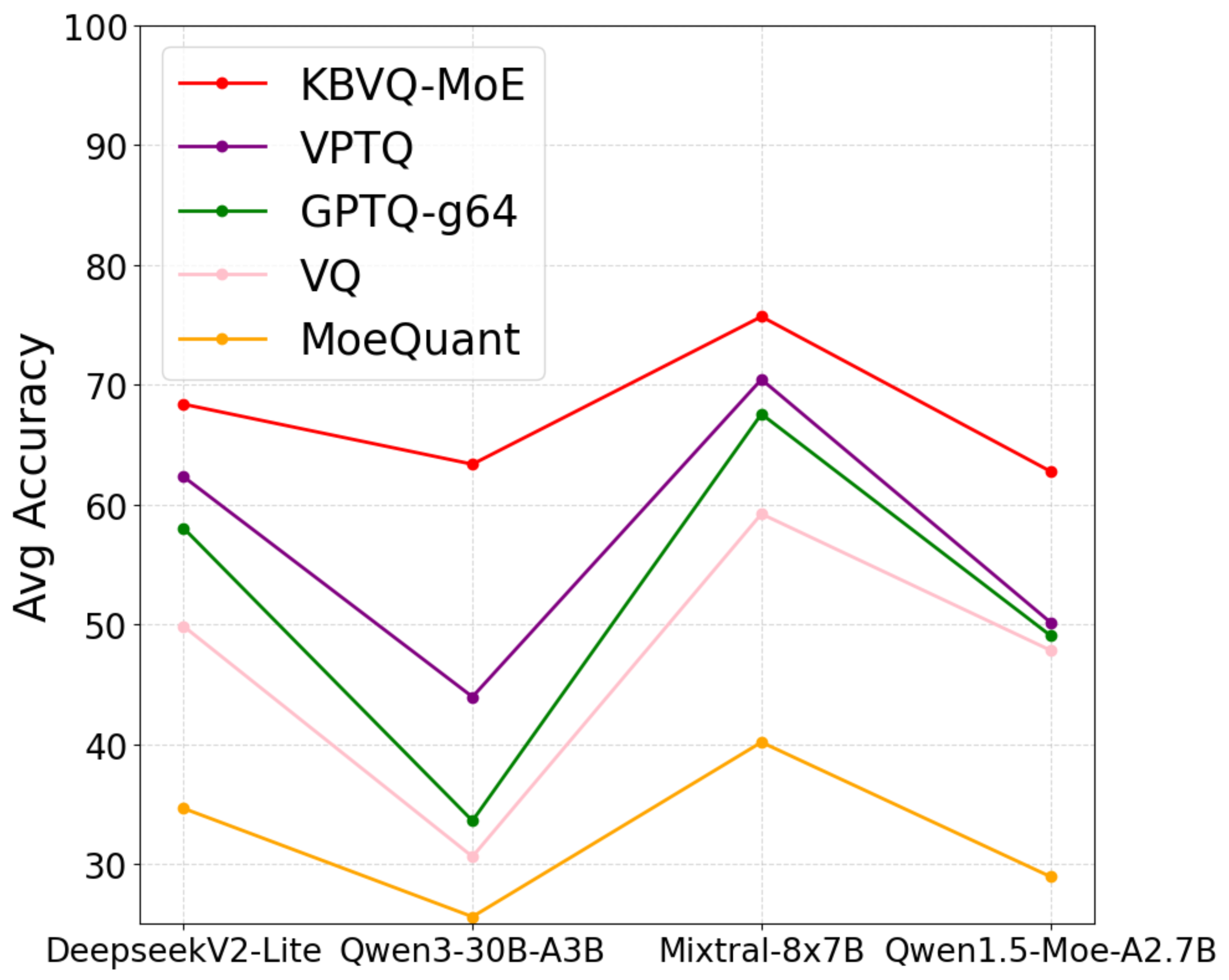}
        \captionof{figure}{
        Average accuracy across multiple MoE architectures, showing that KBVQ-MoE achieves superior performance under 2-bit quantization.
        }
        \label{fig3:motivation_result}
    \end{minipage}\hfill
    \begin{minipage}{0.6\columnwidth}
        \centering
        \includegraphics[width=\linewidth]{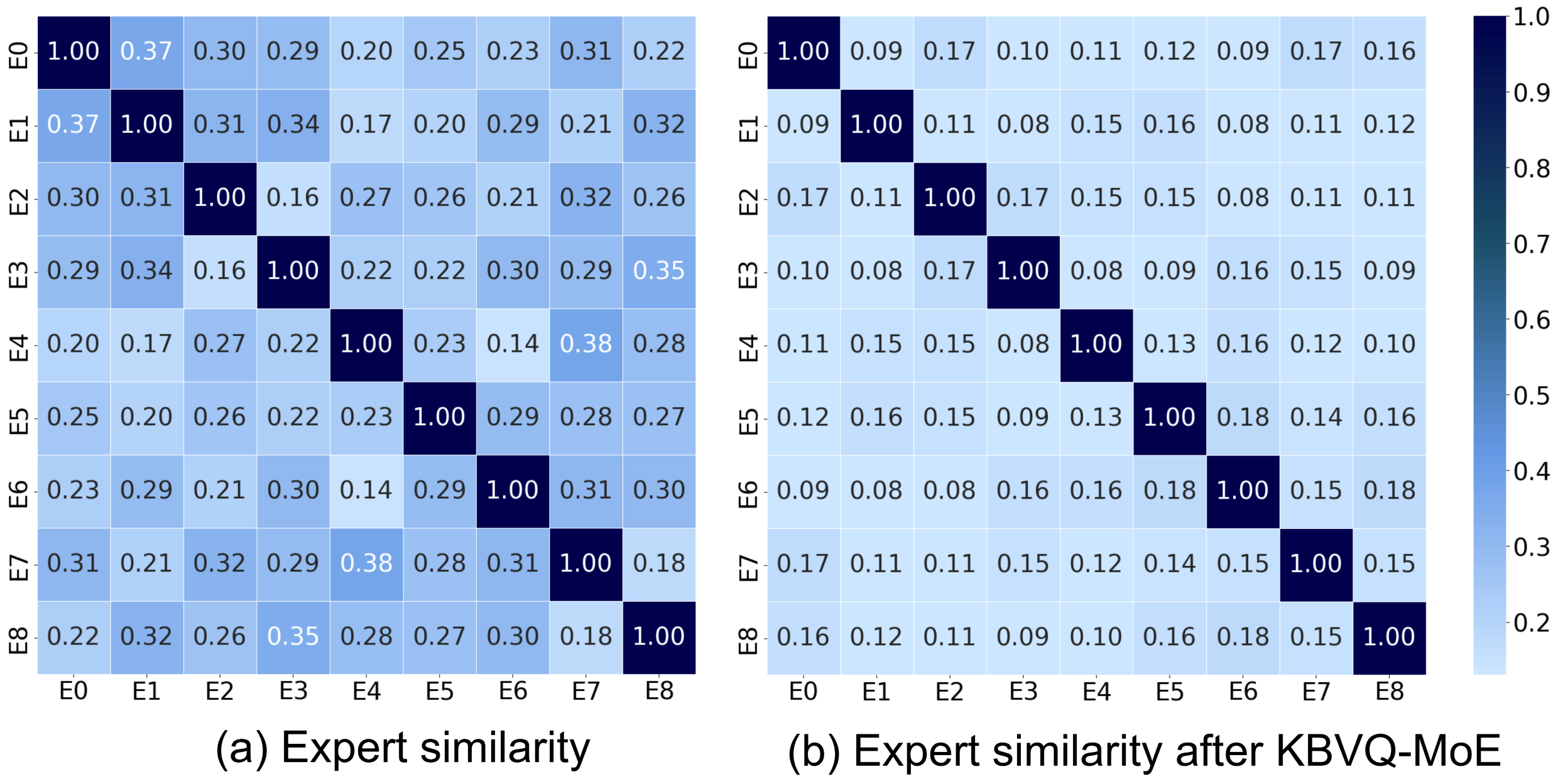}
        \captionof{figure}{
        Similarity of expert outputs before and after redundancy elimination by KBVQ-MoE.
        }
        \label{fig1:motivation_preprocess}
    \end{minipage}
\end{figure}
Post-Training Quantization (PTQ) has emerged as a promising solution for compressing LLMs without the need for retraining \citep{illm, gptq}. 
A typical subcategory of PTQ is Scalar Quantization (SQ), which represents each weight independently by mapping it to a discrete value from a lower bit-width set.
SQ performs well from medium to high bit-width ($\geq 4$ bits) \citep{mambaquant,flatquant,specquant}, but its representational capability drops sharply at extremely low bit-width ($\leq 3$ bits), leading to significant accuracy degradation.
In contrast, Vector Quantization (VQ), another subcategory of PTQ, shows strong potential for ultra-low-bit dense LLM quantization\citep{pcdvq, vptq, qtip}. 
This advantage is realized mainly through leveraging a predefined codebook—where weight vectors are mapped to the most similar discrete codewords within the codebook—thus significantly reducing the data volume while maintaining an acceptable level of information retention.
 

\begin{wrapfigure}{r}{0.5\textwidth}
    \centering
    \includegraphics[width=\linewidth]{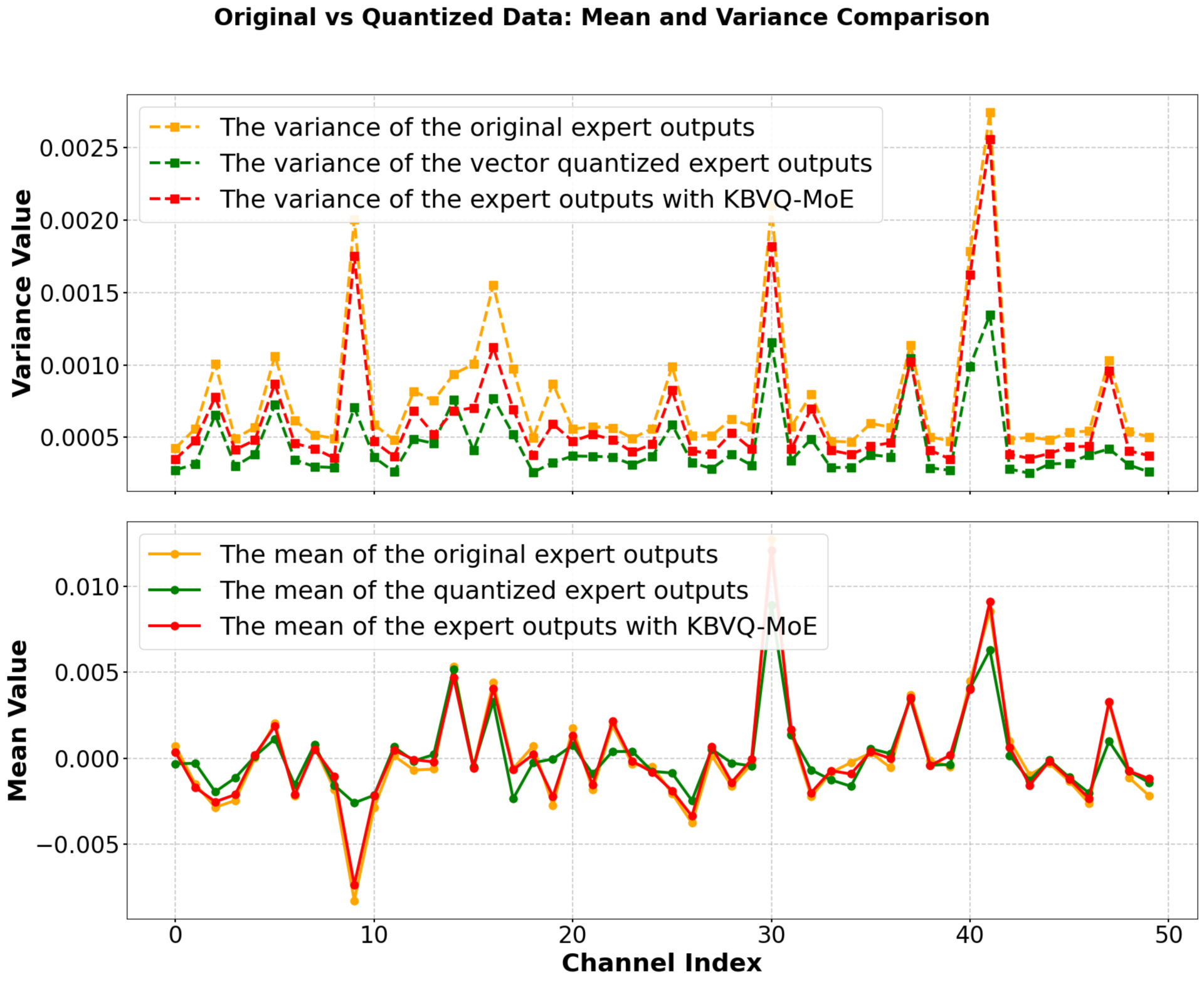}
    \caption{Distributional Shifts in Qwen3-30B-A3B Layer 20 Outputs: (Top) Per-channel Mean Comparisons (FP, Direct VQ, KBVQ-MoE); (Bottom) Per-channel Variance Comparisons (FP, Direct VQ, KBVQ-MoE).}
    \label{fig2:motivation_postprocess}
\end{wrapfigure}
However, directly applying VQ to MoE architectures suffers from significant performance degradation caused by two key obstacles.
\ding{182} \textbf{Redundant representation among experts.} 
MoE experts often capture similar feature patterns\citep {moeadaptation, sankar, load, d2moe, submoe}, resulting in substantial parameter redundancy. As shown in Fig.~\ref{fig1:motivation_preprocess}(a), experts within the same layer produce highly similar outputs for identical activations, reflecting their overlapping functional roles. This redundancy wastes quantization capacity and prevents limited codebook resources from being concentrated on expert-specific (i.e., non-redundant) representations.
\ding{183} \textbf{Cumulative and amplified outputs bias.} 
Quantization errors accumulate across layers, resulting in biased layer outputs. In MoE architectures, this bias becomes more pronounced because expert aggregation further amplifies it, leading to more severe distributional shifts than in dense LLMs. As shown in Fig.\ref{fig2:motivation_postprocess}, both the mean and variance of outputs shift after quantization. When biased outputs from multiple experts are aggregated through gating, the bias is amplified and propagates across layers, leading to distributional drift and degraded model performance.  

To this end, we propose KBVQ-MoE, the first VQ framework tailored to MoE architectures. KBVQ-MoE is built on two efficient innovations:
\ding{182} \textbf{Input-driven redundancy elimination(IDRE)}. 
First, we employ the Karhunen–Loève Transform (KLT) to align expert weights with input activation statistics, thereby mapping them into a common latent space, referred to here as a unified representation (see Eq.~\ref{eq:w_unify}).
Next, we apply SVD to this unified representation in order to extract the dominant shared representation, which are retained at full precision, making the remaining expert-specific representations easier to quantize effectively.
As illustrated in Fig.~\ref{fig1:motivation_preprocess}(b), after redundancy elimination the outputs of experts exhibit much lower similarity compared to Fig.~\ref{fig1:motivation_preprocess}(a), validating the effectiveness of redundancy elimination.
\ding{183} \textbf{Bias-corrected output stabilization(BCOS)}. 
We apply vector quantization only to expert-specific(i.e., non-redundant) representations and stabilize the outputs of quantized experts through lightweight linear scaling and bias correction. 
As a result, with proposed IDRE and BCOS, KBVQ-MoE provide an effective solution for ultra-low-bit quantization in MoE LLMs, achieving both efficient codebook utilization and stable output distributions.

We conducted experimental evaluations of the proposed method on a variety of MoE LLMs, including Qwen1.5-MoE-A2.7B\citep{qwen1_5}, Qwen3-30B-A3B\citep{qwen3}, Mixtral-8x7B\citep{mixtral} and DeepseekV2-Lite\citep{deepseekv2}. As shown in Fig.~\ref{fig3:motivation_result}, KBVQ-MoE consistently outperforms existing scalar and vector quantization methods under the same compression ratio, with particularly strong gains in ultra-low precision settings. For example, at 2-bit quantization on the Qwen3-A3B-30B, our method improves perplexity by 6 and raises average accuracy by nearly 10\%, demonstrating its potential for deploying MoE LLMs on resource-constrained devices such as edge platforms.

The main contributions of this paper are as follows:
\begin{itemize}
    \item We identify two key challenges that arise when vector quantization is applied to MoE-based LLMs: the waste of codebook resources caused by common redundant representation among experts, and outputs bias in quantization errors exacerbated by expert aggregation.
    \item We propose the KBVQ-MoE framework, which integrates Input-driven redundancy elimination and Bias-corrected outputstabilization.
    \item Both theoretical analysis and experimental results demonstrate the effectiveness of the KBVQ-MoE method. It exhibits significant advantages over existing methods on models such as the Qwen series and Mixtral, and even achieves near-floating-point accuracy performance under 2-bit quantization.
\end{itemize}

%% file: sec/3.related_work.tex
\section{Related Work}\label{sec:related_work}
\subsection{Post-Training Quantization (PTQ)}
\paragraph{Scalar Quantization (SQ)} assigns an independent scaling factor and zero point to each tensor (e.g., layer weight matrix), mapping continuous values to discrete integers. Methods such as GPTQ \citep{gptq} and GPTAQ \citep{gptaq} leverage Hessian information to optimize the error propagation path, while I-LLM \citep{illm} achieves quantization of large models through smoothing and full integer inference approximation. Approaches like Quarot \citep{quarot} and OSTQuant \citep{ostquant} incorporate orthogonal transforms to improve quantization efficiency. However, SQ suffers from representational bottlenecks in ultra-low bit-width scenarios ($\leq 4$ bits), making it difficult to balance the compression ratio and accuracy of MoE LLMs.

\paragraph{Vector Quantization (VQ)} clusters weight vectors into shared codebooks and leverages structural redundancy to achieve higher compression ratios \citep{vector_quantize}. GPTVQ \citep{gptvq} combines expectation-maximization with error feedback to optimize codebooks; VPTQ \citep{vptq} and AQLM \citep{aqlm} employ residual quantization to reduce error accumulation; PCDVQ \citep{pcdvq} achieves efficient quantization by decoupling the magnitude and direction of vectors; QuIP\# \citep{quip_sharp} and QTIP \citep{qtip} utilize geometric transformations to improve error distribution. These studies have demonstrated the superiority of VQ at extremely low bit-widths for dense LLMs. Nevertheless, when existing VQ techniques are directly applied to MoE LLMs, they fail to account for the unique structural information of MoE, resulting in suboptimal compression performance.

\subsection{MoE LLMs Compression Methods}
Research on model compression for MoE architectures is still in its infancy. Most methods directly adopt general model compression techniques, lacking special consideration for the structural characteristics of MoE gating and multi-expert architectures. EAC-MoE \citep{eacmoe} reduces parameters by pruning redundant experts, but pruning causes irreversible loss of functionality. D2-MoE \citep{d2moe} and SubMoe \citep{submoe} decompose expert weights into low-rank factors, yet their compression ratios are limited by rank constraints, failing to meet the demands of extreme resource-constrained scenarios. MoEQuant \citep{moequant} uses routing statistics to balance the contributions of each expert during calibration, but its performance remains unsatisfactory under quantization of $\leq 4$ bits.

In summary, existing MoE LLMs compression methods lack a collaborative optimization mechanism that integrates the statistical characteristics of input activations—they neither fully exploit input-related common patterns shared across experts nor specifically correct distribution shifts caused by quantization errors of experts. This makes it difficult for them to balance model accuracy at high compression ratios. The KBVQ-MoE framework proposed in this paper is designed to fill this gap, significantly improving the quantization performance of MoE LLMs through input-driven redundancy elimination and bias-corrected output stabilization.

%% file: sec/4.preliminaries.tex
\section{Preliminaries}
\label{sec:Preliminaries}

\paragraph{Mixture-of-Experts.} 
Mixture-of-Experts (MoE) architectures modify standard Transformer layers by replacing conventional feed-forward (MLP) modules with specialized MoE modules, enabling efficient scaling of model capacity while maintaining computational efficiency\citep{6797059,6796382}. MoE layer typically comprises two types of components: a set of $m$ shared experts ($\{E^s_1, \ldots, E^s_m\}$), a set of $n$ routing experts ($\{E^r_1, \ldots, E^r_n\}$), and a gating network (or router network) that determines which experts process each input token.  

Each expert—whether shared or routing—functions as a lightweight feed-forward sub-network (effectively a compact MLP), with shared experts designed to handle general patterns across inputs and routing experts specialized for specific input subsets. When processing an input hidden state $\bm{x} \in \mathbb{R}^d$, the MoE layer operates through a structured sequence:  

1. Gating Score Calculation: The gating network computes weights (or affinities) $g_i(\bm{x})$ for all experts, quantifying how well each expert aligns with the input $\bm{x}$. These weights reflect the probability of routing $\bm{x}$ to the corresponding expert.  

2. Expert Selection: For routing experts, only the top $k$ experts with the highest affinities are selected, denoted by the subset $\mathcal{K} = \mathrm{top}k\left(\{g_j(\boldsymbol{x})\mid j\in\{1,\ldots,n\}\}\right)$. In contrast, all shared experts are typically utilized to preserve general representational capacity.  

3. Weighted Output Computation: The final output $\bm{y}$ of the MoE layer is a weighted sum of outputs from the selected experts, where the weights are the corresponding gating scores. Mathematically, this is formalized as:  

\begin{equation}
    \begin{aligned}
    \label{eq:output_moe}
    \bm{y} = \underbrace{\sum_{i=1}^{m} E^s_i(\bm{x}) g_i(\bm{x})}_{\text{shared experts}} + \underbrace{\sum_{j\in\mathcal{K}} E^r_j(\bm{x}) g_j(\bm{x})}_{\mathrm{top~}k\text{ routing experts}},
    \end{aligned}
\end{equation}  

While this formulation captures the core mechanism, specific designs vary across architectures—for example, in how gating scores are computed, how experts are structured (e.g., depth, activation functions), or how the balance between shared and routing experts is tuned.

%% file: sec/5.method.tex
\section{Method}\label{sec:method}
This section introduces the proposed KLT-guided SVD with Bias-Corrected Vector Quantization for MoE Large Language Models (KBVQ-MoE). 
The framework is built on two innovations that align with the challenges identified in Section~\ref{sec:introduction}:  
(1) \textbf{Input-driven Redundancy Elimination (IDRE)}, where the Karhunen–Loève Transform (KLT) maps expert weights into a unified representation aligned with input statistics, and singular value decomposition (SVD) is then applied to extract dominant shared components retained at full precision, leaving expert-specific (i.e., non-redundant) representations that are more amenable to quantization;  
(2) \textbf{Bias-Corrected Output Stabilization (BCOS)}, where vector quantization is applied only to the expert-specific representations, and lightweight linear scaling with bias correction is used to stabilize outputs and mitigate distributional shifts.  
The overall workflow of KBVQ-MoE is mathematically formulated in Eq.\ref{eq1:pre-post-process}, while a detailed algorithmic flow is provided in Appendix~\ref{alg:kbvq_moe}.
\begin{equation} \label{eq1:pre-post-process}
    W  \xrightarrow[\text{KLT+SVD}]{\text{IDRE}} \underbrace{W_{\text{share}}}_{\text{Shared Part}}  +  \underbrace{W_{\text{quant}}}_{\text{Specific Part}} \xrightarrow[\text{Bias Correction}]{\text{BCOS}}\underbrace{W_{\text{share}}}_{\text{no quant}}  +  \underbrace{W_{\text{quant}}}_{\text{vector quant}}+\underbrace{(s,b)}_{\text{scale\&bias}.}
\end{equation}

\subsection{Input-driven redundancy elimination(IDRE)} \label{sec:method1}
In MoE architectures, experts often exhibit substantial redundancy, as their weights encode overlapping mappings for identical activations. 
This redundancy wastes quantization capacity and prevents the limited codebook from focusing on expert-specific representations. 
To address this issue, we introduce \textbf{Input-driven Redundancy Elimination (IDRE)}, which leverages the statistical characteristics of model inputs to construct a unified representation of expert weights. 
By isolating shared components and retaining them at full precision, IDRE reduces redundancy and improves the efficiency of codebook utilization for subsequent quantization. 
IDRE is applied uniformly to the MLP weights of all experts in each MoE layer—including both shared and routing experts. 
The detailed transformation of the MoE structure before and after applying IDRE is illustrated in Fig.~\ref{fig4:method1}. 
This procedure consists of the following steps:

\begin{figure}[!h]
    \centering
    \includegraphics[width=0.75\linewidth]{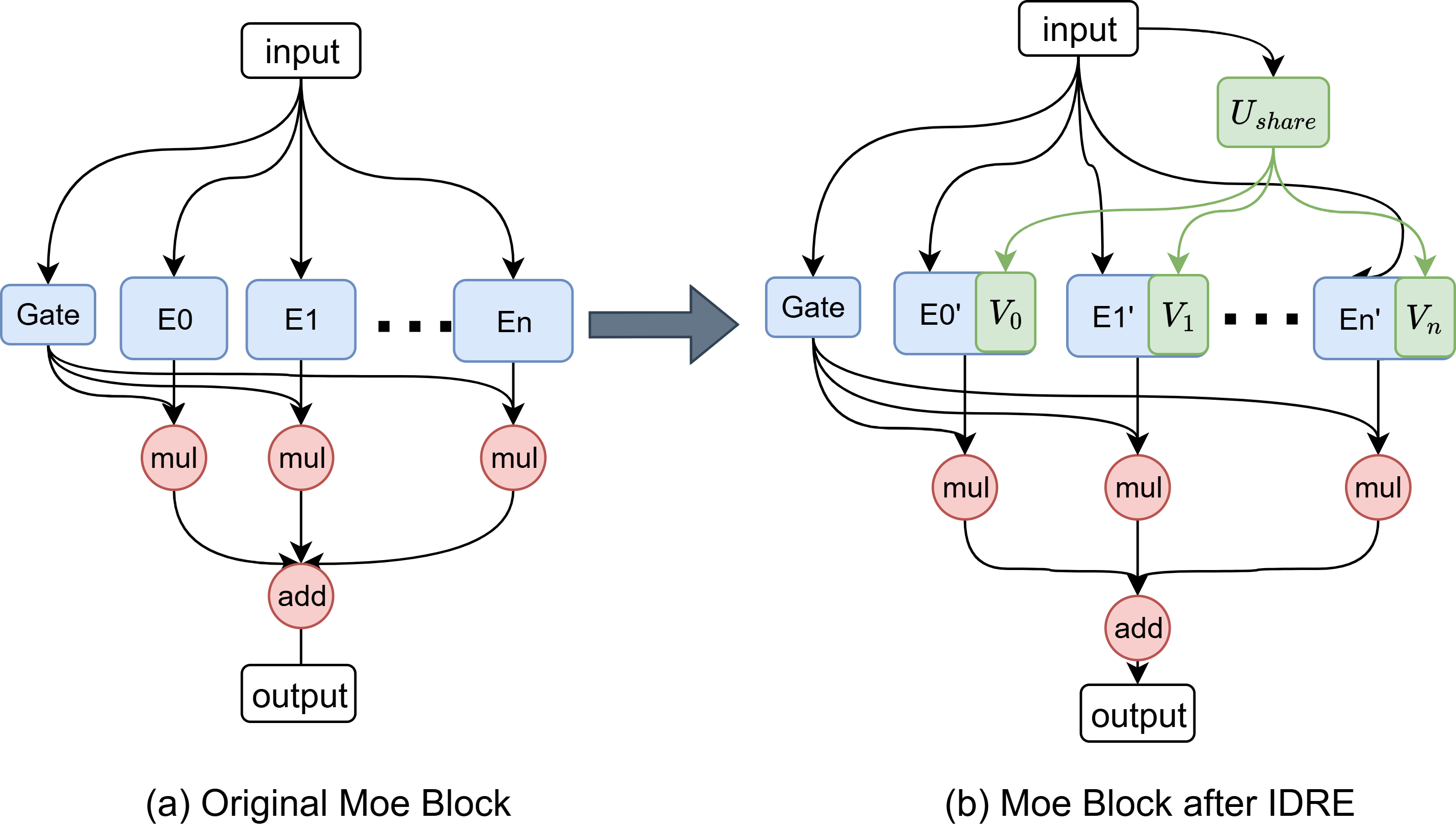}
    \caption{The comparison of structural changes in the MoE (Mixture-of-Experts) structure before and after IDRE.}
    \label{fig4:method1}
\end{figure}
\paragraph{Step 1: KLT Decomposition of Input Activations: Constructing the Input Coherence Space.}
To reduce redundancy, we begin by applying the Karhunen–Loève Transform (KLT) to the input activations in order to construct a coherence basis that captures their dominant directions. 
Let the input activation matrix of the expert layer be \(X \in \mathbb{R}^{b \times ic}\), where \(b\) denotes the number of samples and \(ic\) is the input dimension. 
The procedure is as follows:
\begin{enumerate}
    \item The covariance matrix is computed as \(C_X = \frac{1}{B-1} X^T X \in \mathbb{R}^{ic \times ic}.\)
    \item An eigenvalue decomposition is performed on \(C_X\):\(C_X = \left(U_{\text{KLT}} \Lambda_{\text{KLT}}^{\frac{1}{2}}\right)^T \left(U_{\text{KLT}} \Lambda_{\text{KLT}}^{\frac{1}{2}}\right)\). where \(U_{\text{KLT}} = [u_1, \dots, u_{ic}]\) is an orthonormal basis, and \(\Lambda_{\text{KLT}} = \mathrm{diag}(\lambda_1 \ge \dots \ge \lambda_{ic})\) is a diagonal matrix of eigenvalues. Here, \(u_j\) represents the j-th coherence direction of the input, and \(\lambda_j\) denotes the energy magnitude of this direction.
    \item Based on this, we construct the input coherence basis: \(U_X = U_{\text{KLT}} \Lambda_{\text{KLT}}^{1/2}\)
    The column vectors of the resulting \(U_X\) are sorted by input energy magnitude, forming an orthogonal coordinate system based on input statistics. This facilitates prioritizing the extraction of common weight structures related to high-energy input directions in subsequent steps.
\end{enumerate}
A detailed theoretical derivation linking this decomposition to the minimization of quantization error is provided in Appendix~\ref{appendix:methd1_klt2mse}.

\paragraph{Step 2: Mapping Weights to the Input Coherence Space.} 
To ensure that weight analysis is guided by the dominant directions of the input, we project the original weight matrix \(W \in \mathbb{R}^{oc \times ic}\) onto the input coherence basis obtained in Step 1:  \(\hat{W} = W U_{X} \in \mathbb{R}^{oc \times ic},\) where $oc$ denotes the output channel dimension and $ic$ denotes the input channel dimension.
In this transformed representation, each column of \(\hat{W}\) corresponds to the mapping of a specific coherence direction of the input to the output space. 
The \(j\)-th column, in particular, represents how the weights project along the \(j\)-th most energetic input direction \(u_j\).  
This alignment establishes a direct connection between weight structures and input characteristics, which facilitates the subsequent extraction of shared and expert-specific components.  

\paragraph{Step 3: Extracting Input-Coherent Common Structures: Precise Redundancy Elimination.} The common structures of MoE experts manifest as cross-expert shared weight patterns in the input coherence space (e.g., similar mappings for the high-energy direction \(u_1\)). To capture these patterns, we concateniate the mapped weights of all $n$ experts($\hat{W}^{(i)}$ denotes the i-th
expert)—along their output channel dimension—into a $(n \cdot oc)\times ic$-dimensional unified representation:
\begin{equation}\label{eq:w_unify}
    \bar{W} = \begin{bmatrix} \hat{W}^{(1)} \\ \hat{W}^{(2)} \\ \vdots \\ \hat{W}^{(n)} \end{bmatrix} = \begin{bmatrix} W^{(1)} U_{\text{X}} \\ W^{(2)} U_{\text{X}} \\ \vdots \\ W^{(n)} U_{\text{X}} \end{bmatrix} \in \mathbb{R}^{(n \cdot oc) \times ic}
\end{equation}
We then apply SVD to \(\bar{W}\) in order to extract the dominant shared representation:\(\bar{W} = \left(U \Sigma V^T\right)^T.\)
\begin{enumerate}
    \item \(U \in \mathbb{R}^{ic \times k}\) denotes the shared left singular vectors (shared mapping directions across experts, where $k$ is the number of retained dominant singular values, focusing on input-coherent common patterns); \(\Sigma = \text{diag}(\sigma_1, \dots, \sigma_k)\) is the singular value matrix (the larger \(\sigma_j\) is, the more prominent the corresponding dominant shared representation).
    \item The common structures in the input coherence space are mapped back to the original weight space via the inverse transformation of the input coherence basis (\(U_{\text{X}}^{-1}\)): 
    \begin{equation}
        U_{\text{share}} = U^T \cdot U_{\text{X}}^{-1} \in \mathbb{R}^{k \times ic}.
    \end{equation}
    \(V \in \mathbb{R}^{(n \cdot oc) \times k}\) represents the right singular vectors, which are partitioned by expert into:
    \begin{equation}
        V = \left[\Sigma_k V^{(1)^T}, \Sigma_k V^{(2)^T}, \dots, \Sigma_k V^{(n)^T}\right]^T,
    \end{equation}
    where \(V^{(i)} \in \mathbb{R}^{oc \times k}\) is the private right singular vector for expert $i$.
    \item The truncated low-rank components \(U_{\text{share}}\) and \(V\) obtained from SVD are retained in full precision to preserve the fidelity of shared and expert-specific representations.
\end{enumerate}

Thus, IDRE explicitly decouples shared structures (kept at full precision) from expert-specific representations (subject to quantization), enabling more efficient use of codebook capacity. 
A rigorous spectral characterization of this decomposition, including its optimality and error bounds, is provided in Appendix~\ref{appendix:methd1_all}.

\subsection{Bias-corrected output stabilization(BCOS)}\label{sec:method2}
While IDRE reduces redundancy and improves codebook utilization, quantization of the remaining expert-specific weights \(W_{quant}\) still introduces cumulative bias that distorts output distributions. 
To mitigate this problem, BCOS stabilizes the outputs through vector quantization followed by lightweight channel-wise affine compensation.

\paragraph{Step 1: Vector Quantization of Expert-Specific Weights.} 
We perform VQ quantization on the remaining expert-specific weights \(W_{\text{quant}}\) in the original space: partition \(W_{\text{quant}}\) into subvectors \(z \in \mathbb{R}^d\), map these subvectors to indices\(q = \arg\min_j \|z - c_j\|^2\) via a codebook \(C = \{c_1, ..., c_K\}\), and the quantized result is \(z_q = c_q\). The quantized matrix is denoted as:
\(W_{\text{quant,VQ}} \in \mathbb{R}^{oc \times ic}\). The compressed weight is the sum of the shared component and the quantized component:
\begin{equation}
    W_{\text{VQ}} = W_{\text{share}} + W_{\text{quant,VQ}}    
\end{equation}
The quantization error, arising solely from the quantization of the specific component, is defined as \(\epsilon = W_{\text{VQ}} - W\).

\paragraph{Step 2: Channel-Wise Bias Correction.} 
As the number of layers increases, the quantization error \(\epsilon\) leads to distributional shifts in the outputs. we apply channel-wise affine compensation to align the statistics of quantized outputs with those of the original model. The corrected output of the linear layer is formulated as
\begin{equation}
    \mathbf{y}_{\text{corr}} = W_{\text{VQ}} x + s \odot (W_{\text{VQ}} x) + b = (s+1) \odot (W_{\text{VQ}} x) + b,
\end{equation}
where \(s \in \mathbb{R}^{oc}\) is a channel-wise scaling factor; \(b \in \mathbb{R}^{oc}\) is a channel-wise bias term; \(\odot\) denotes element-wise multiplication. Here, \(W_{\text{VQ}} x\) is the original quantized output. We choose \(s\) and \(b\) to ensure that the corrected output \(\mathbf{y}_{\text{corr}}\) matches the mean and variance of the full-precision output \(\mathbf{y} = W \mathbf{x}\). Following unbiased estimation in the sense of Minimum Mean Square Error (MMSE) (see Appendix~\ref{appendic:bias_correct}), the optimal parameters are approximated as
\begin{equation}
    s_j \approx \frac{\sigma_{y_j}}{\sigma_{\hat{y}_j}} - 1, 
    \qquad 
    b_j = \mu_{y_j} - (1+s_j) \mu_{\hat{y}_j}.
\end{equation}
where \(\sigma\) and \(\mu\) denote the standard deviation and mean, respectively, and both operations are channel-wise. In other words, each channel is scaled by the ratio of standard deviations between the original and quantized outputs, and shifted by the corresponding mean difference. After this correction, both the mean and variance of each channel are aligned with the full-precision baseline, effectively eliminating the distributional shift caused by quantization. This correction introduces only $2oc$ additional parameters per layer and requires only lightweight per-channel scaling and shifting operations during inference, resulting in negligible computational and storage overhead.

%% file: sec/6.experiments.tex
\section{Experiment}
\paragraph{Models and Datasets.} 
To comprehensively verify the performance effectiveness and scenario applicability of the KBVQ-MoE framework, we conducted experiments on multiple sets of representative Mixture-of-Experts (MoE) models and diverse evaluation tasks.  
For the selection of evaluation models, we prioritized currently mainstream pre-trained MoE LLMs to ensure the reference value of the experimental results. 
Specifically, these models include the Qwen series (e.g., Qwen1.5-MoE-A2.7B \citep{qwen1_5}, Qwen3-30B-A3B \citep{qwen3}), DeepseekV2-Lite \citep{deepseekv2}, and Mixtral \citep{mixtral}.
In terms of the design of evaluation tasks and datasets, we focused on natural language reasoning and understanding benchmarks, and comprehensively examined the framework’s performance through multi-dimensional metrics:  
- On the one hand, we tested the perplexity (ppl) of the language model (LLM) output with a sequence length of 4096 on the WikiText2 dataset \citep{wikitext2}, to measure the coherence of language generation and semantic modeling capability of the compressed model;  
- On the other hand, we evaluated the model accuracy on 7 zero-shot datasets, including Arc-Challenge \citep{arc-c}, Arc-Easy \citep{arc-e}, HellaSwag \citep{hellaswag}, LAMBADA-openai \citep{lambda}, LAMBADA-standard \citep{lambda}, PIQA \citep{piqa}, and WinoGrande \citep{winogrande}. This was done to comprehensively verify the generalization capability of the compressed model across different task scenarios.

\paragraph{Baselines.}  
To demonstrate the effectiveness and superiority of our method, we compare KBVQ-MoE with a range of well-validated excellent methods, such as Round To Nearest (RTN) and GPTQ \citep{gptq}, the MoEQuant \citep{moequant} (proposed specifically for MoE quantization), as well as VQ. We respectively compare the accuracy differences of these methods under ultra-low bit-widths (2\~3 bits), with their compression ratios evaluated based on actual storage usage.
\input{tab/table1}
\paragraph{Implementation Details.}  
All experiments were conducted on NVIDIA RTX A6000 GPU. The calibration dataset used in our experiments was sampled from the Red\_Pajama dataset \citep{redpajama}: we fixed the random seed to 42 and randomly sampled 256 data samples with a sequence length of 4096 from the Red\_Pajama dataset, which served as the calibration set for the IDRE and BCOS methods in this paper. In the IDRE process, we set the truncated rank$k$ of KLT-SVD to \(1/128\) of the full rank,  and under this configuration, the average number of parameters increased by SVD redundancy extraction is only 0.12. In this paper, the length of the vector quantization vector length is set to 4. This selection will reduce the occupation of the codebook to a low level. Meanwhile, the k-means algorithm adopted in the vector quantization process uses kmeans++ for initialization and sets the iteration number to 100, which helps balance the stability of codebook generation and the efficiency of quantization computation. All evaluations were performed using the open-source LM-Evaluation-Harness toolkit \citep{lm-eval}.

\paragraph{Main results.}   
Table~\ref{tab1:main} presents the results of direct comparisons between our method and other approaches. Overall, our method demonstrates significant advantages across different models and quantization bit-widths, achieving average accuracy (Avg Acc) close to full precision (FP16) while maintaining low error.  
Particularly in low-bit quantization scenarios ($\leq 3$ bits), our method exhibits notable robustness and generalization capability. For Qwen models, take Qwen3-30B-A3B\citep{qwen3} under 2-bit quantization: its PPL decreases by nearly 6 points, and Avg Acc increases by more than 10 points. For Qwen1.5-MoE-A2.7B\citep{qwen1_5} and Mixtral-8x7B\citep{mixtral} under 3-bit quantization, their Avg Acc reaches 67.99 and 78.35, respectively, which are nearly identical to the FP16 precision. Detailed accuracy results for each dataset and model are provided in Appendix~\ref{appendix:all_expriments}

\paragraph{Ablation Experiments.} 
\input{tab/table2}
As shown on Table~\ref{tab2:ablation1}. In the 3-bit quantization of Qwen3-30B-A3B, our complete scheme (IDRE + BCOS) achieves the best performance across the board: the perplexity (ppl) on Wikitext2 decreases to 9.26, representing a 50.5\% reduction compared to the unprocessed baseline (18.72) — this is a 20.6\% decrease compared to the IDRE-only scheme and a 35.3\% decrease compared to the BCOS-only scheme. Additionally, comprehensive improvements are observed across five benchmark tasks.
From the ablation of individual modules, it is evident that IDRE is the primary source of performance gain, while BCOS further corrects residuals and unleashes a synergistic effect on this foundation. Their combination enables a significant leap in overall accuracy, demonstrating that the proposed two innovations can approach full-precision performance under extremely low bit-widths. This validates its strong robustness and practical value for real-world deployment.

\input{tab/table3}
In IDRE, we use the Karhunen–Loève Transform (KLT) to extract the input coherence basis for redundancy removal across experts, which enables more direct redundancy extraction. In Table~\ref{tab3:ablation2}, we compare the direct performance between redundancy extraction without input coherence and that with KLT-based input coherence. Clearly, the IDRE scheme yields greater accuracy improvement: specifically, on the Qwen1.5-MoE-A2.7B model, the perplexity (ppl) on Wikitext2 decreases by more than 2 points.

\begin{minipage}[t]{0.47\textwidth}
For the IDRE, we conducted ablation studies to determine the optimal ratio of the truncated rank $k$ of SVD to the full rank $n$. Table~\ref{tab4:ablation3} presents the ablation results of Qwen3-30B-A3B under 2-bit quantization: without IDRE, the model performs poorly, with a perplexity (ppl) of 15.3 on Wikitext2. However, after applying SVD truncation to pre-extract expert redundancy, the ppl drops sharply to 11.87. Further increasing 
\end{minipage}
\hfill
\begin{minipage}[t]{0.50\textwidth}
\vspace{-0.6cm}
\input{tab/table4}
\vspace{-0.3cm}
\end{minipage}
the truncated rank, performance improvements become marginal, while the average bit-width increases instead. Therefore, we recommend setting the truncated rank to \(1/128\) of the full rank.

\begin{minipage}[t]{0.49\textwidth}
To further validate the effectiveness of our approach on MoE architectures, we integrate our IDRE and BCOS modules with different vector quantization baselines, treating them as a general plug-in component for MoE quantization. This design demonstrates that our method is both versatile and efficient, as it can be seamlessly combined with existing quantization techniques to enhance their performance. As shown in Table~\ref{tab10:vq_method}, on Qwen1.5-MoE-A2.7B\citep{qwen1_5} applying our method to GPTVQ\citep{gptvq} under 3-bit quantization yields nearly a 30\% ppl improvement, while under 2-bit quantization with VPTQ\citep{vptq}, our method still achieves more than a 15\% performance gain.
\end{minipage}
\hfill
\begin{minipage}[t]{0.48\textwidth}
\vspace{-0.6cm}
\input{tab/table10}
\vspace{-0.3cm}
\end{minipage}

\begin{minipage}[t]{0.29\textwidth}
We conduct a speed evaluation in the decoder stage. In our tests with 1k input tokens, as shown in Table~\ref{tab9:speedup}, the Qwen1.5-MoE-A2.7B model under 2-bit quantization (integrated with KBVQ
\end{minipage}
\hfill
\begin{minipage}[t]{0.70\textwidth}
\vspace{-0.6cm}
\input{tab/table9}
\vspace{-0.3cm}
\end{minipage}
-MoE and corresponding VQ methods) achieves an inference speedup of over 1.5$\times$, further validating the practical deployment value of the proposed framework in balancing accuracy retention and efficiency.

%% file: tab/table1.tex
\begin{table}[!h]
\caption{Comparison of average accuracy between the KBVQ-MoE method and other quantization methods on various different moe models}
\label{tab1:main}
\resizebox{1.0\textwidth}{!}{
\begin{tabular}{c|c|cc|cc|cc|cc}
\hline
 &             & \multicolumn{2}{c|}{Qwen1.5-MoE-A2.7B}  & \multicolumn{2}{c|}{Qwen3-30B-A3B}       & \multicolumn{2}{c|}{Mixtral-8x7B}       & \multicolumn{2}{c}{DeepseekV2-Lite}     \\ \cline{3-10} 
\multirow{-2}{*}{Bit} & \multirow{-2}{*}{Method}         & W2 (↓)  & Avg Acc (↑)                   & W2 (↓)   & Avg Acc (↑)                   & W2 (↓)  & Avg Acc (↑)                   & W2 (↓)  & Avg Acc (↑)                   \\ \hline
16 & FP16        & 7.22    & 68.07    & 8.70     & 70.24    & 3.88    & 78.57    & 5.92    & 70.68    \\ \hline
 & RTN         & 638509  & 25.64    & 765922   & 25.89    & 274952  & 25.27    & 174653  & 25.12    \\
 & GPTQ        & 12.69   & 49.07    & 438.42   & 33.06    & 5.69    & 67.56    & 8.49    & 58.04    \\
 & MoeQuant    & 583542  & 34.64    & 37465    & 28.94    & 13.43   & 40.16    & 25893   & 25.59    \\
 & VQ          & 26.95   & 47.84    & 115.30   & 30.61    & 5.99    & 59.22    & 10.96   & 49.85    \\
\multirow{-5}{*}{2}   & \cellcolor[HTML]{F2F3F5}KBVQ-MoE & \cellcolor[HTML]{F2F3F5}\textbf{9.61} & \cellcolor[HTML]{F2F3F5}\textbf{62.78} & \cellcolor[HTML]{F2F3F5}\textbf{11.87} & \cellcolor[HTML]{F2F3F5}\textbf{63.37} & \cellcolor[HTML]{F2F3F5}\textbf{5.39} & \cellcolor[HTML]{F2F3F5}\textbf{75.69} & \cellcolor[HTML]{F2F3F5}\textbf{7.94} & \cellcolor[HTML]{F2F3F5}\textbf{63.10} \\ \hline
 & RTN         & 116689  & 25.68    & 68657    & 25.84    & 45136   & 25.80    & 97.75   & 33.32    \\
 & GPTQ        & \textbf{7.58}    & 66.36    & 11.32    & 63.51    & 4.17    & 77.43    & 6.98    & \textbf{68.89}    \\
 & MoeQuant    & 7.87    & 66.79    & 24.85    & 37.32    & 5.45    & 72.21    & 7.52    & 66.34    \\
 & VQ          & 11.47   & 55.94    & 18.72    & 52.06    & 5.52    & 73.98    & 7.94    & 62.26    \\
\multirow{-5}{*}{3}   & \cellcolor[HTML]{F2F3F5}KBVQ-MoE & \cellcolor[HTML]{F2F3F5}7.74 & \cellcolor[HTML]{F2F3F5}\textbf{67.99} & \cellcolor[HTML]{F2F3F5}\textbf{9.26}  & \cellcolor[HTML]{F2F3F5}\textbf{69.09} & \cellcolor[HTML]{F2F3F5}\textbf{4.07} & \cellcolor[HTML]{F2F3F5}\textbf{78.35} & \cellcolor[HTML]{F2F3F5}\textbf{6.95} & \cellcolor[HTML]{F2F3F5}68.73 \\ \hline
\end{tabular}
}
\end{table}

%% file: tab/table2.tex
\begin{table}[h!]
\caption{Ablation studies on the pre-process and post-process of KBVQ-MoE respectively}
\label{tab2:ablation1}
\resizebox{1.0\textwidth}{!}{
\begin{tabular}{c|c|ll|llllll}
\hline
\multicolumn{1}{l|}{Model}     & \multicolumn{1}{l|}{Bit}  & IDRE & BCOS & W2   & ARC-E & ARC-C & HE    & PIQA  & WI    \\ \hline
\multirow{5}{*}{Qwen3-30B-A3B} & \multicolumn{1}{l|}{FP16} & -           & -            & 8.70 & 79.25 & 56.40 & 59.60 & 80.30 & 70.48 \\ \cline{2-10} 
 & \multirow{4}{*}{3} & \ding{55} & \ding{55} & 18.72 & 57.83 & 40.87 & 63.23 & 71.82 & 57.70 \\
 &                    & \checkmark & \ding{55} & 11.67 & 71.35 & 50.55 & 73.51 & 77.75 & 66.92 \\
 &                    & \ding{55} & \checkmark & 14.32 & 65.49 & 47.33 & 68.37 & 75.52 & 60.42 \\
 &                    & \checkmark & \checkmark & \textbf{9.26 } & \textbf{77.27} & \textbf{53.24} & \textbf{75.53 }& \textbf{78.89} & \textbf{70.01} \\ \hline
\end{tabular}
}
\end{table}

%% file: tab/table3.tex
\begin{table}[h!]
\caption{Ablation on using KLT to guide SVD for expert redundancy}
\label{tab3:ablation2}
\resizebox{1.0\textwidth}{!}{
\begin{tabular}{l|l|l|llllll}
\hline
Model                              & Bit                & Method     & W2    & ARC-E & ARC-C & HE    & PIQA  & WI    \\ \hline
\multirow{2}{*}{Qwen1.5-MoE-A2.7B} & \multirow{2}{*}{2} & SVD+VQ     & 14.03 & 68.21 & 43.59 & 67.32 & 76.60 & 66.80 \\
                                   &                    & KLT-SVD+VQ & 11.87 & 70.33 & 47.61 & 67.49 & 76.66 & 66.46 \\ \hline
\end{tabular}
}
\end{table}

%% file: tab/table4.tex
\begin{table}[H]
\centering  
\caption{Ablation study on the rank of SVD truncation in the Pre-Process procedure}
\vspace{-3mm}
\label{tab4:ablation3}
\begin{tabular}{l|l|l|l}
\hline
Model                          & Bit  & k/n   & W2     \\ \hline
\multirow{4}{*}{Qwen3-30B-A3B} & 2.01 & 0.0   & 15.30 \\
                               & 2.08 & 1/128 & 11.87  \\
                               & 2.11 & 1/64  & 11.30  \\
                               & 2.20 & 1/32  & 11.01  \\ \hline
\end{tabular}
\end{table}

%% file: tab/table10.tex
\begin{table}[H]
\centering
\caption{Experimental Results of Different VQ Methods Integrated with KBVQ-MoE for Qwen1.5-Moe-A2.7B Model Under 2-bit Quantization}
\vspace{-3mm}
\label{tab10:vq_method}
\begin{tabular}{l|ll|l}
\hline
Base Method            & IDRE & BCOS  & W2    \\ \hline
\multirow{4}{*}{GPTVQ} & \ding{55} & \ding{55} & 12.88 \\
                       & \checkmark  & \ding{55} & 11.03 \\
                       & \ding{55} & \checkmark  & 11.97 \\
                       & \checkmark  & \checkmark  & 9.43  \\ \hline
\multirow{4}{*}{VPTQ}  & \ding{55} & \ding{55} & 10.17 \\
                       & \checkmark  & \ding{55} & 9.21  \\
                       & \ding{55} & \checkmark  & 9.77  \\
                       & \checkmark  & \checkmark  & 8.78  \\ \hline
\end{tabular}
\end{table}

%% file: tab/table9.tex
\begin{table}[H]
\centering
\caption{Decoder speed test}
\vspace{-3mm}
\label{tab9:speedup}
\begin{tabular}{l|lll}
\hline
\multirow{2}{*}{Model} & \multicolumn{3}{l}{Decoder Speed (Tokens/s)} \\ \cline{2-4} 
                       & BF16        & Quantized      & Speed Up      \\ \hline
Qwen1.5-Moe-A2.7B      & 22.31       & 35.24          & 1.58          \\
Qwen3-30B-A3B          & 10.85       & 17.37          & 1.60          \\ \hline
\end{tabular}
\end{table}

%% file: sec/7.conclusion.tex
\vspace{-0.3cm}
\section{Conclusion}
\vspace{-0.3cm}
This paper proposes KBVQ-MoE, the first vector quantization framework tailored to Mixture-of-Experts (MoE) architectures. 
We address two critical challenges of applying vector quantization to MoE LLMs: (1) redundant representation among  experts, which wastes limited codebook capacity, and (2) cumulative and amplified outputs bias, which manifests during expert aggregation and leads to severe distributional shifts.
KBVQ-MoE integrates two key techniques: \textbf{Input-driven Redundancy Elimination (IDRE)}, which isolates shared structures and retains them at full precision, and \textbf{Bias-Corrected Output Stabilization (BCOS)}, which applies vector quantization only to expert-specific representations while correcting output distribution shifts through channel-wise affine compensation. 
Both theoretical analysis and empirical evaluations on diverse MoE LLMs (e.g., Qwen and Mixtral) demonstrate that KBVQ-MoE achieves superior performance under ultra-low-bit quantization, preserving accuracy with negligible runtime overhead. 
In future work, we will extend the framework to adaptive codebook design and expert pruning, further improving the deployment efficiency of large-scale MoE models in resource-constrained environments.

%% file: sec/9.appendix.tex
\clearpage
\appendix
\section{Appendix}

\subsection{Use of LLMs}
Large language model (LLM) tools were employed solely for text polishing, including refinement of sentence structure, lexical optimization, and enhancement of language fluency.  

All core components of this research—such as the conception of research ideas and framework, algorithm design and implementation, experimental protocol design, data collection and processing, execution and analysis of experiments, and validation of conclusions—were independently completed by the research team.  

LLM tools were not involved in the conception of research content, generation of technical solutions, execution of experiments, or derivation of conclusions. The authors affirm full responsibility for the scientific validity, authenticity, and originality of the work in accordance with academic integrity standards.

\subsection{Theoretical Proof of the Input Coherence Basis
}\label{appendix:methd1_klt2mse}
Let the linear mapping of the i-th expert be \(y^{(i)} = W^{(i)}x\), with its quantized version denoted as \(\tilde{W}^{(i)}\) and the error matrix defined as \(E^{(i)} \triangleq \tilde{W}^{(i)} - W^{(i)}\). For the input covariance \(C_X = \mathbb{E}[xx^T]\) that reflects the true distribution during inference, a natural definition of task distortion is the output mean squared error (MSE):
\begin{equation}
    \mathcal{L} \triangleq \sum_{i=1}^n \mathbb{E}\left[\left\| \tilde{W}^{(i)}x - W^{(i)}x \right\|_2^2\right] = \sum_{i=1}^n \mathrm{Tr}\!\left(E^{(i)} C_X E^{(i)T}\right).
\end{equation}
This metric directly measures the impact of quantization on the output side, rather than the unweighted difference of the weights themselves. Perform the Karhunen–Loève Transform (KLT) on \(C_X\), such that \(C_X = U_{\mathrm{KLT}} \Lambda_{\mathrm{KLT}} U_{\mathrm{KLT}}^T\), where \(\Lambda_{\mathrm{KLT}} = \mathrm{diag}(\lambda_1 \ge \dots \ge \lambda_{ic})\). Define the input coherence basis as \(U_X \triangleq U_{\mathrm{KLT}} \Lambda_{\mathrm{KLT}}^{1/2}\). Then the following identity holds:
\begin{equation}
    \mathrm{Tr}(E C_X E^T) = \mathrm{Tr}\!\left(E U_X (U_X)^T E^T\right) = \|E U_X\|_F^2,
\end{equation}
thus
\begin{equation}
    \mathcal{L} = \sum_{i=1}^n \left\| (\tilde{W}^{(i)} - W^{(i)}) U_X \right\|_F^2.
\end{equation}
This indicates that: in the input coherence space, the output MSE is equivalent to the Frobenius norm of the weight error. Therefore, all "extraction/quantization" operations aimed at reducing output distortion should be performed in this space. Denote \(\hat{W}^{(i)} \triangleq W^{(i)} U_X\).

\subsection{Spectral characterization of the optimal shared subspace (KLT–SVD).}
\label{appendix:methd1_all}
To formalize the redundancy removal performed by IDRE, we stack all expert weights along the output dimension:
\begin{equation}
  W \,\triangleq\,
  \begin{bmatrix}
    \hat{W}^{(1)} \\
    \vdots \\
    \hat{W}^{(n)}
  \end{bmatrix}
  \in \mathbb{R}^{(n\!\cdot\!o_c)\times i_c},
  \qquad
  S \,\triangleq\, W^\top W.
\end{equation}
Here $\hat{W}^{(i)}$ denotes the $i$-th expert weight expressed in the input KLT coordinates introduced in Appendix.~\ref{appendix:methd1_klt2mse}. The matrix $S$ is the (input–aligned) Gram matrix of all experts.
In practice, IDRE computes a truncated SVD of the stacked matrix $W \in \mathbb{R}^{(n\cdot o_c)\times i_c}$: $W = U \Sigma V^\top$. 
Note that the Gram matrix$S = W^\top W$ admits the eigen-decomposition $S = V \Sigma^2 V^\top$, i.e., the eigenvalues of $S$ are the squared singular values of $W$, and the eigenvectors of $S$ coincide with the right singular vectors of $W$. 
Therefore, selecting the top–$k$ eigenvectors of $S$ is exactly equivalent to performing a rank–$k$ truncated SVD on $W$ and keeping the top–$k$ right singular vectors.
The parameter $k$ used in our analysis and implementation is the dimension of this shared right-singular subspace.

\paragraph{Optimal shared subspace.}
We seek a $k$-dimensional shared subspace that maximizes the stacked energy of all experts:
\begin{equation}
  \max_{U \in \mathbb{R}^{i_c \times k},\; U^\top U = I}
    \mathrm{Tr}\!\big(U^\top S U\big)
  \;=\;
  \max_{U^\top U = I} \big\|W U\big\|_F^2.
\end{equation}
By Ky Fan's theorem, the optimizer $U_k$ is given by the top–$k$ eigenvectors of $S$ associated with its largest eigenvalues $\{\sigma_j^2\}_{j=1}^k$. Let $P_k \triangleq U_k U_k^\top$ be the projector onto this shared subspace. In the KLT space, each expert admits the orthogonal decomposition
\begin{equation}
  \check{W}^{(i)}
  \;=\;
  \underbrace{\check{W}^{(i)} P_k}_{\text{shared (redundancy)}}
  \;+\;
  \underbrace{\check{W}^{(i)} (I - P_k)}_{\text{specific (difference)}} ,
\end{equation}
and mapping back to the original coordinates yields $W_{\text{share}}^{(i)} = \hat{W}^{(i)} P_k T^{-1}$ and $W_{\text{spec}}^{(i)} = W^{(i)} - W_{\text{share}}^{(i)}$, where $T$ is the KLT transform defined in Sec.~A.2. Intuitively, the shared component captures directions that are simultaneously important for many experts, while the specific component collects expert–unique variations.
\paragraph{Why KLT before SVD.}
Let $x$ be the input to the MoE layer with covariance $\Sigma_X = \mathbb{E}[x x^\top] = U_X \Lambda_X U_X^\top$ and define the input KLT transform by $T = U_X \Lambda_X^{1/2}$. In this basis, the stacked Gram matrix takes the form
\begin{equation}
  S
  = W^\top W
  = T^\top \Big(\sum_{i=1}^n W^{(i)\top} W^{(i)}\Big) T
  = \Lambda_X^{1/2} U_X^\top
    \Big(\sum_{i=1}^n W^{(i)\top} W^{(i)}\Big)
    U_X \Lambda_X^{1/2}.
\end{equation}
Thus the spectrum $\{\sigma_j^2\}$ of $S$ jointly reflects \emph{both} the input energy (through $\Lambda_X$) and the across–expert weight energy (through $\sum_i W^{(i)\top}W^{(i)}$) along each input principal direction. Selecting the top–$k$ eigenvectors of $S$ therefore retains those directions that are simultaneously dominant in the input distribution and heavily utilized across experts, which is
precisely the notion of ``shared'' structure we aim to preserve in full precision.
\paragraph{Truncation error and redundancy removal.}
The residual (expert–specific) part after projecting onto $P_k$ satisfies
\begin{align}\label{eq:truncation-error-16}
  \sum_{i=1}^n \big\|\check{W}^{(i)} (I - P_k)\big\|_F^2
  \;=\;
  \big\|W (I - P_k)\big\|_F^2
  \;=\;
  \sum_{j>k} \sigma_j^2,
\end{align}
where $\{\sigma_j^2\}_{j=1}^{i_c}$ are the eigenvalues of $S$ sorted in decreasing order. Hence the truncation error is \emph{exactly} controlled by the tail eigenvalues discarded by $P_k$. Correspondingly, the redundancy removal ratio of IDRE at rank $k$ can be quantified as
\begin{equation}
  \rho_k
  \;\triangleq\;
  \frac{\sum_{j=1}^k \sigma_j^2}{\sum_{j=1}^{i_c} \sigma_j^2},
\end{equation}
which measures the fraction of stacked expert energy captured by the shared subspace.

\begin{figure}[!h]
    \centering
    \includegraphics[width=\linewidth]{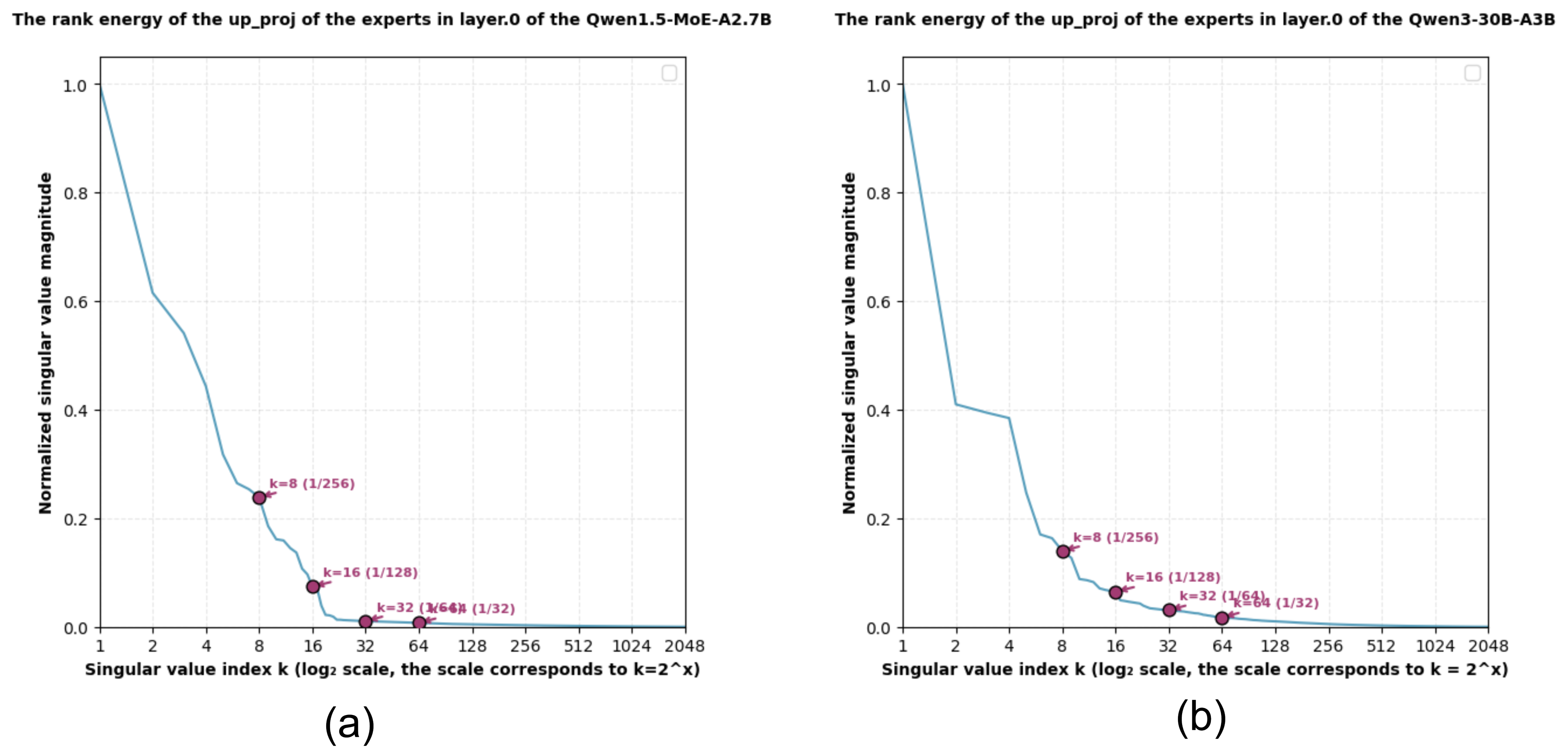}
    \caption{The low-rank characteristics after expert fusion activation of the first block, (a) Qwen1.5-MoE-A2.7B, (b) Qwen3-30B-A3B}
    \label{fig5:singular_value}
\end{figure}

This characterization provides an explicit trade-off: increasing $k$ monotonically raises $\rho_k$ and reduces the truncation error in~\eqref{eq:truncation-error-16}, but also enlarges the full-precision shared component. 
Empirically, for MoE LLMs such as Qwen1.5-MoE-A2.7B and Qwen3-30B-A3B, we observe that the spectrum of $S$ is strongly low-rank and decays rapidly (approximately heavy–tailed). As shown in the Fig.\ref{fig5:singular_value}, it demonstrates the obvious low-rank characteristics of the expert groups and their posterior singularity.. 
Under a simple power-law approximation $\sigma_j^2 \propto j^{-\alpha}$ with $\alpha>1$, the residual fraction $1-\rho_k$ behaves like $C k^{1-\alpha}$; once $k$ exceeds a moderate threshold, the marginal gain from further increasing $k$ quickly diminishes.
Across layers and models, choosing $k \approx i_c/128$ typically yields $\rho_k \approx 0.6$–$0.8$, while larger ranks (e.g., $k = i_c/64$) provide only marginal perplexity improvements at noticeably higher storage cost.Tab.\ref{tab:low_rank} shows a comparison of k selection for two moe models of Qwen, and the low rank of 1/128 can already achieve very good results.
\begin{table}[h]
\centering
\caption{Low-rank value ablation}
\label{tab:low_rank}
\begin{tabular}{l|lllll}
\hline
k                 & 1/256 & 1/128 & 1/64  & 1/48  & 1/32  \\ \hline
Compress Ratio    & 2.05  & 2.08  & 2.11  & 2.15  & 2.2   \\ \hline
Qwen1.5-MoE-A2.7B & 20.71 & 9.61  & 9.60  & 9.58  & 9.58  \\ \hline
Qwen3-30B-A3B     & 15.30 & 11.87 & 11.30 & 11.30 & 11.01 \\ \hline
\end{tabular}
\end{table}
This explains why $k = i_c/128$ serves as a robust operating point that balances reconstruction
error (controlled by the tail eigenvalues) and the overhead of the shared full-precision subspace across different models and tasks.

\subsection{System Bias Correction} \label{appendic:bias_correct}
In the paper, we reformulate bias correction as a residual affine transform on the quantized output:
\begin{equation}
    y_{\text{corr}} = W_{\text{VQ}} x + s \odot (W_{\text{VQ}} x) + b 
    = (1+s) \odot (W_{\text{VQ}} x) + b,
\end{equation}
where $s, b \in \mathbb{R}^{oc}$ denote channel-wise residual scaling and bias, respectively.  
We now show that this correction is statistically optimal under the mean squared error (MSE) criterion.

\paragraph{1. Problem Setup}  
Let the full-precision output be $y = W x$ and the quantized output be $\hat{y} = W_{\text{VQ}} x$.  
Our goal is to determine $s$ and $b$ such that
\[
y_{\text{corr}} = (1+s)\odot \hat{y} + b
\]
approximates $y$ as closely as possible.  
The optimization problem can be written as
\begin{equation}
    (s^*, b^*) = \arg\min_{s, b} \; \mathbb{E}\!\left[\|y - ((1+s)\odot \hat{y} + b)\|^2\right].
\end{equation}

\paragraph{2. Closed-form Solution}  
For the $j$-th output channel, this reduces to:
\begin{equation}
    (s_j^*, b_j^*) = \arg\min_{s_j, b_j} \; \mathbb{E}\!\left[(y_j - ((1+s_j)\hat{y}_j + b_j))^2\right].
\end{equation}
This is equivalent to a univariate linear regression with slope $\alpha_j = 1+s_j$ and intercept $b_j$.  
Its closed-form solution is
\begin{equation}
    \alpha_j^* = \frac{\mathrm{Cov}(y_j, \hat{y}_j)}{\mathrm{Var}(\hat{y}_j)}, 
    \qquad 
    b_j^* = \mu_{y_j} - \alpha_j^* \mu_{\hat{y}_j},
\end{equation}
and therefore
\begin{equation}
    s_j^* = \alpha_j^* - 1.
\end{equation}

\paragraph{3. Approximation under High Correlation}  
Because $y_j$ and $\hat{y}_j$ differ only by quantization noise, they are highly correlated.  
Thus, we approximate
\begin{equation}
    \mathrm{Cov}(y_j, \hat{y}_j) \approx \sigma_{y_j}\,\sigma_{\hat{y}_j}.
\end{equation}
Substituting into the regression solution gives
\begin{equation}
    s_j^* \approx \frac{\sigma_{y_j}}{\sigma_{\hat{y}_j}} - 1, 
    \qquad 
    b_j^* = \mu_{y_j} - (1+s_j^*) \mu_{\hat{y}_j}.
\end{equation}

\paragraph{4. Conclusion}  
This yields exactly the mean–variance matching rule used in the main text, showing that the residual affine correction is not a heuristic adjustment but an \emph{unbiased MMSE-optimal estimator}.

\subsection{Compression Condition}\label{appendix:Compression_Condition}
In the current experiment, all methods are matched by weight bit-width (2 bits or 3 bits), while the additional overhead of KBVQ-MoE (shared low-rank components and per-channel biases) only slightly increases the average number of bits per weight (by approximately 0.08 bits). In the table\ref{tab:Compression_Condition}, we report the actual storage usage and effective bits per parameter of all baselines, and provide comparisons under equal compression ratios.
\begin{table}[h]
\caption{ compression achieved by various techniques }
\label{tab:Compression_Condition}
\resizebox{\textwidth}{!}{
\begin{tabular}{l|l|c|cccc}
\hline
Bit  & Method   & \multicolumn{1}{l|}{Compress ratio} & \multicolumn{1}{l}{Qwen1.5-MoE-A2.7B} & \multicolumn{1}{l}{Qwen3-30B-A3B} & \multicolumn{1}{l}{Mixtral-8x7B} & \multicolumn{1}{l}{DeepseekV2-Lite} \\ \hline
16   & FP16     & 0\%                                 & 27.9GB                                & 59.1GB                            & 89.6GB                           & 29.3GB                              \\
2    & RTN      & 87.5\%                              & 4.3GB                                 & 8.4GB                             & 12.6GB                           & 4.3GB                               \\
2.25 & GPTQ     & 85.9\%                              & 4.7GB                                 & 9.3GB                             & 13.5GB                           & 4.7GB                               \\
2    & MoeQuant & 87.5\%                              & 4.3GB                                 & 8.4GB                             & 12.6GB                           & 4.3GB                               \\
2.08 & VQ       & 87\%                                & 4.3GB                                 & 8.7GB                             & 13GB                             & 4.4GB                               \\
2.08 & KBVQ-MoE & 87\%                                & 4.3GB                                 & 8.7GB                             & 13GB                             & 4.4GB                               \\ \hline
\end{tabular}
}
\end{table}

We provide a comprehensive quantification from  \textbf{three perspectives} : offline computation overhead, inference-time overhead, and memory/parameter cost. These analyses complement the “Decoder speed test” in Table 6 in manuscript and present a complete view of the efficiency characteristics of KBVQ-MoE.

(1) Offline Computation Overhead (Does Not Affect Inference Latency) 

KBVQ-MoE introduces an additional one-time  KLT-guided SVD decomposition  during the calibration phase. This operation is applied after stacking all expert weights: 
The cost of KLT-SVD is  comparable to a single forward pass  on the calibration set. 
It occurs  entirely offline  during calibration. 
It introduces  no additional inference-time latency  or compute cost.
Therefore, the additional computation introduced by KBVQ-MoE does  not  affect actual inference performance.

(2) Inference-Time Overhead (Nearly Negligible) 

During inference, KBVQ-MoE introduces only:
Two extra parameters per output channel  (scale and bias). A lightweight per-channel affine correction  after each expert’s linear layer: $y_{\text{corr}} = (1 + s) \odot (W_{\text{VQ}} x) + b$. 
This involves only element-wise additions and multiplications, whose complexity is  far lower  than the original MatVec operation inside each expert MLP.
Empirically, this affine correction accounts for  less than 0.1\% of the expert forward FLOPs , consistent with the  1.5–1.6× inference speedup  reported in Table 6.
Thus, KBVQ-MoE introduces  minimal additional compute overhead during inference.

(3) Detailed Memory and Parameter Overhead Analysis 

Let:
 Expert weight dimension: $m \times l$,
 Number of experts: n,
 IDRE rank ratio: k,
 VQ subvector length: v,
 Quantization bit-width: b.

The total bit consumption of KBVQ-MoE consists of four parts:
 Original FP16 parameters:$16nml$,
 IDRE shared low-rank components:$16 (m + ln)\min(m,l)k$,
 VQ codebook + indices:$mlbn + 16 \cdot 2^{bv} \cdot v \cdot n$,
 BCOS scale/bias (per channel):$16 \cdot 2 \cdot ln$,
 Total compression ratio:$\frac{16(m+ln)\min(m,l)k + mlbn + 2^{(bv+4)}vn + 32ln}{16nml}$.

Example: Qwen1.5-MoE-2.7B (gate\_proj weight) ,
 Weight dimension: $5632 \times 2048$,
 Number of experts: 64,
 Original storage:$5632 \times 2048 \times 64 \times 16; \text{bit}
  = 1.38;\text{GB}$,
 KBVQ-MoE configuration:
   2-bit VQ,
   vector length = 4,
   $k = \frac{1}{128}$,
Total compressed storage: $((5632 + 2048\times 64)\cdot 2048/128 \cdot 16) + (5632 \cdot 2048 \cdot 2 \cdot 64 + 2^{8} \cdot 4 \cdot 16 \cdot 64) + (2048\cdot 2 \cdot 16 \cdot 64),\text{bit} = 0.18,\text{GB}$.

This corresponds to an  87\% compression ratio , equivalent to  2.08 effective bits per weight.

Through the analyses above, we provide a complete efficiency characterization of KBVQ-MoE across:
Offline computation:  acceptable one-time cost, no inference impact. 
Inference computation:  negligible additional overhead (<0.1\% FLOPs),
Memory cost:  clear and formula-based quantification (87\% compression demonstrated).

\subsection{4bit ablation experiment}\label{appendix:4bit_exp}
Our primary focus in this work is to address the challenges of ultra–low-bit compression (e.g., 2–3 bits), where existing methods often suffer from severe accuracy degradation. In higher-bit regimes, standard scalar quantization (SQ) can already achieve strong performance, and the relative advantage of vector quantization (VQ) gradually diminishes. Moreover, our method is based on vector quantization, whose codebook training via $k$-means clustering incurs a computational cost that grows exponentially with the target bit-width, making VQ less attractive in practice for high-bit settings. For completeness, we additionally report the performance of KBVQ-MoE at 4-bit in Tab.\ref{tab:4bit_exp}, which shows that our method remains competitive in this regime, even though ultra–low-bit compression is the main target of our design.
\begin{table}[h]
\centering
\caption{4bit ablation experiment}
\label{tab:4bit_exp}
\resizebox{0.8\textwidth}{!}{
\begin{tabular}{c|l|lllll}
\hline
\multicolumn{1}{l|}{Model}         & Method   & W2    & ARC-C & ARC-E & PIQA  & WI    \\ \hline
\multirow{6}{*}{Qwen1.5-MoE-A2.7B} & FP16     & 7.22  & 69.53 & 44.03 & 80.47 & 69.30 \\
                                   & RTN      & 10.83 & 64.95 & 41.12 & 76.84 & 67.28 \\
                                   & GPTQ     & 7.43  & 68.03 & 43.78 & 78.90 & 68.14 \\
                                   & MoeQuant & 7.55  & 68.97 & 44.04 & 79.76 & 67.86 \\
                                   & VQ       & 8.92  & 64.68 & 40.88 & 77.20 & 64.91 \\
                                   & KBVQ-MoE & 7.59  & 69.05 & 44.39 & 80.33 & 68.56 \\ \hline
\multirow{6}{*}{Mixtral-8x7B}      & FP16     & 3.84  & 85.39 & 66.38 & 85.20 & 76.72 \\
                                   & RTN      & 5.41  & 75.84 & 55.48 & 77.84 & 70.73 \\
                                   & GPTQ     & 4.03  & 84.89 & 63.78 & 84.21 & 74.97 \\
                                   & MoeQuant & 4.12  & 83.88 & 63.70 & 84.20 & 75.02 \\
                                   & VQ       & 4.27  & 84.18 & 64.21 & 84.93 & 74.81 \\
                                   & KBVQ-MoE & 4.16  & 84.97 & 65.04 & 84.84 & 74.17 \\ \hline
\end{tabular}
}
\end{table}

\clearpage
\subsection{All Experiments}\label{appendix:all_expriments}
\input{tab/table5}
\input{tab/table6}
\input{tab/table7}
\input{tab/table8}

\input{algo/algo}

\subsection{Ablation of IDRE and BCOS}\label{appendix:ablation_of_IDRE_and_BCOS}
To further clarify the contribution of each component within the IDRE and BCOS modules, we conduct fine-grained, component-wise ablations at 2-bit on Qwen1.5-MoE-A2.7B. This directly addresses the reviewer’s concern regarding the lack of quantitative analysis for the internal steps of each module.

(a) Effect of the KLT step in IDRE.
IDRE consists of (i) an input-driven KLT transform and (ii) a subsequent SVD on the stacked expert weights in the KLT space. To isolate the effect of the KLT step, we compare the full IDRE against a variant that performs SVD directly on the original stacked weights (without KLT alignment). The results at 2-bit are summarized in Tab.\ref{tab:ablation_of_KLT_in_IDRE}.
\begin{table}[h]
\centering
\caption{Ablation of KLT in IDRE on Qwen1.5-MoE-A2.7B (2-bit).}
\label{tab:ablation_of_KLT_in_IDRE}
\resizebox{0.8\textwidth}{!}{
\begin{tabular}{c|l|lll}
\hline
\multicolumn{1}{l|}{Model}         & IDRE    & W2            & HE             & WI             \\ \hline
\multirow{2}{*}{Qwen1.5-MoE-A2.7B} & w/o KLT & 11.77         & 35.89          & 63.01          \\
                                   & KLT     & \textbf{9.61} & \textbf{39.59} & \textbf{65.11} \\ \hline 
\end{tabular}
}
\end{table}

(b) Effect of mean and variance correction in BCOS.
BCOS performs channel-wise bias correction using both the mean and variance of the channel outputs. To disentangle their respective roles, we evaluate four variants: (i) no correction, (ii) mean-only, (iii) variance-only, and (iv) full mean+variance correction. The 2-bit results on Qwen1.5-MoE-A2.7B are given in Tab.\ref{tab:ablation_of_mean_var_in_BCOS}.
\begin{table}[h]
\centering
\caption{Ablation of mean and variance correction in BCOS on Qwen1.5-MoE-A2.7B (2-bit).}
\label{tab:ablation_of_mean_var_in_BCOS}
\resizebox{0.8\textwidth}{!}{
\begin{tabular}{l|l|l|lll}
\hline
Model                              & Mean & Variance & W2            & HE             & WI             \\ \hline
\multirow{4}{*}{Qwen1.5-MoE-A2.7B} & ×    & ×        & 11.03         & 35.18          & 63.49          \\
                                   & \checkmark    & ×        & 11.01         & 35.35          & 63.40          \\
                                   & ×    & \checkmark        & 10.38         & 36.70          & 64.23          \\
                                   & \checkmark    & \checkmark        & \textbf{9.61} & \textbf{39.59} & \textbf{65.11} \\ \hline
\end{tabular}
}
\end{table}

\subsection{Comparison of MoE compression methods}\label{appendix: MoE_compression_methods}
The core designs of D2-MoE, SubMoE, and EAC-MoE all revolve around "expert layer structure optimization", which belongs to a different technical branch of MoE compression compared to KBVQ-MoE's "weight quantization". The specific differences can be summarized as follows:
\begin{itemize}
    \item SubMoE: Merges similar experts through subspace clustering and extracts cross-expert shared low-rank features via SVD to reduce the number of experts. Essentially, it is a structural compression method combining "expert-level pruning + low-rank decomposition". The compression ratio is mainly determined by the expert merging ratio (e.g., 50\% means merging half of the experts).
    \item D2-MoE: Decomposes expert weights into a "shared base matrix + expert-specific delta matrix" and performs low-rank approximation on the delta matrix to reduce the number of parameters. Essentially, it is a structural compression method combining "weight decomposition + low-rank approximation". The compression ratio is determined by the rank truncation ratio of the delta matrix (e.g., retaining 60\% of the rank).
    \item EAC-MoE: Starts from "expert selection behavior" and jointly models two operations: quantization and pruning. It proposes QESC (Quantization with Expert-Selection Calibration), which explicitly calibrates the router during low-bit quantization to mitigate expert selection bias caused by quantization noise. It also proposes PESF (Pruning based on Expert-Selection Frequency), which prunes "cold experts" using their actual selection frequency to further improve inference speed and reduce memory usage while minimizing accuracy loss. Thus, EAC-MoE can be regarded as a "routing-aware quantization + expert pruning hybrid compression scheme".
    \item KBVQ-MoE (ours): Does not change the number or structure of experts. First, it extracts cross-expert shared low-rank subspaces in the input KLT space via IDRE to explicitly remove redundancy, then performs vector quantization on expert-specific residuals. Essentially, it is an "MoE structure-aware weight quantization method". The compression ratio is determined by the quantization bit-width (2–3 bits) and codebook reuse efficiency, achieving 80\%–90\% memory compression under the same structural constraints.
\end{itemize}

Overall, D2-MoE/SubMoE are more inclined to directly reduce the number of parameters (removing experts / reducing rank), EAC-MoE follows a hybrid route of structural pruning + quantization, while KBVQ-MoE focuses on achieving high compression ratios through structure-aware low-bit quantization while preserving the original expert structure. Different routes may exhibit distinct trade-offs in terms of "expert specificity preservation", "inference speed", and "implementation complexity", which need to be characterized through empirical comparisons.

\paragraph{Fair Comparison Experiments: Settings and Results.}
To ensure fairness, we uniformly use Mixtral-8×7B as the baseline model (retaining the original expert structure: 8 experts, 32 layers) and adopt identical experimental settings:
\begin{itemize}
    \item Evaluation tasks: WikiText2 (language modeling, metric: perplexity (PPL), lower is better), ARC-Challenge (ARC\_C), ARC-Easy (ARC\_E), WinoGrande (WinG) (all zero-shot inference, metric: accuracy, higher is better);
    \item Compression ratio definition: Memory saving is uniformly calculated as "the ratio of compressed parameter storage to the original FP16 model" to avoid misunderstandings caused by inconsistent definitions of "compression ratio" across different methods;
    \item Experimental environment: NVIDIA RTX A100 GPU, PyTorch 2.1, evaluation tool: LM-Evaluation-Harness (v0.4.0).
\end{itemize}

The comparison results are shown in the Tab.\ref{tab:moe_compress_compare} (Note: ARC\_C, ARC\_E, and WinG values are accuracy, retained to two decimal places for precision):

\begin{table}[h]
    \centering
    \caption{Ablation KBVQ—MoE and other MoE compression methods}
    \label{tab:moe_compress_compare}
    \begin{tabular}{lccccc}
        \hline
        Method          & Memory saving & WikiText2 (PPL) & ARC\_C & ARC\_E & WinG \\
        \hline
        Sub-MoE         & 50\%          & 6.97            & 0.45  & 0.75  & 0.72 \\
        D2-MoE          & 60\%          & 6.46            & 0.38  & 0.72  & 0.71 \\
        EAC-MoE         & 84\%          & 4.58            & 0.55  & 0.81  & 0.75 \\
        KBVQ-MoE (ours) & 87\%          & 4.07            & 0.63  & 0.85  & 0.76 \\
        \hline
    \end{tabular}
\end{table}

In the experiments on Mixtral-8×7B, significant differences exist among the three compression approaches: 
KBVQ-MoE (quantization compression) achieves an 87\% memory saving, which is significantly higher than that of Sub-MoE (50\%) and D2-MoE (60\%), and slightly higher than that of EAC-MoE (84\%). 
Moreover, it performs optimally across all tasks: its Perplexity (PPL) on WikiText2 is 2.39-2.90 lower than that of Sub-MoE/D2-MoE and 0.51 lower than that of EAC-MoE; its accuracy on ARC\_C is 0.18-0.25 higher than that of Sub-MoE/D2-MoE and 0.08 higher than that of EAC-MoE. 
These results confirm that KBVQ-MoE can well retain the expressive ability of MoE experts under a high compression ratio (>80\%).

EAC-MoE (hybrid compression), which combines routing calibration and expert pruning based on selection frequency, outperforms Sub-MoE and D2-MoE (structural compression) in performance. 
However, KBVQ-MoE still shows better performance under a similar compression ratio and does not require modifying the expert structure. 
Additionally, KBVQ-MoE preserves the original expert topology, achieving an accuracy of 0.63 on complex reasoning tasks (e.g., ARC\_C)—far exceeding the 0.38-0.45 of Sub-MoE/D2-MoE and the 0.55 of EAC-MoE. 
It also features greater flexibility in deployment and migration: it can be directly adapted to various MoE LLMs (e.g., Qwen, DeepSeek) with only one round of offline calibration and codebook training, demonstrating prominent comprehensive advantages.

\subsection{challenging benchmarks}\label{appendix: challenging_benchmarks}
We carried out ablation studies between our method and diverse baselines on more arduous tasks shown on Tab.\ref{tab:challenging_benchmarks}. Main tasks include: MMLU\citep{mmlu}, MathQA\citep{mathqa}, GSM8K\citep{gsm8k}, HumanEval\citep{humaneval}.
\begin{table}[h]
\centering
\caption{more challenging benchmarks}
\label{tab:challenging_benchmarks}
\resizebox{0.8\textwidth}{!}{
\begin{tabular}{c|c|ccccc}
\hline
Model                              & Bits               & Method                        & MMLU  & MathQA & GSM8K & HumanEval \\ \hline
\multirow{5}{*}{Qwen1.5-MoE-A2.7B} & 16                 & \multicolumn{1}{c|}{FP16}     & 59.60 & 37.55  & 62.55 & 32.32     \\ \cline{2-7} 
                                   & \multirow{4}{*}{2} & \multicolumn{1}{c|}{GPTQ}     & 26.94 & 19.33  & 15.42 & 13.88     \\
                                   &                    & \multicolumn{1}{c|}{MoeQuant} & 34.75 & 22.42  & 23.21 & 18.12     \\
                                   &                    & \multicolumn{1}{c|}{VQ}       & 48.79 & 28.34  & 46.51 & 28.03     \\
                                   &                    & \multicolumn{1}{c|}{KBVQ-MoE} & 52.16 & 30.75  & 56.39 & 30.11     \\ \hline
\multirow{5}{*}{Qwen3-30B-A3B}     & 16                 & \multicolumn{1}{c|}{FP16}     & 79.5  & 58.53  & 85.44 & 71.5      \\ \cline{2-5} \cline{7-7} 
                                   & \multirow{4}{*}{2} & \multicolumn{1}{c|}{GPTQ}     & 42.11 & 34.9   & 49.34 & 50.12     \\
                                   &                    & \multicolumn{1}{c|}{MoeQuant} & 45.1  & 39.86  & 57.38 & 55.32     \\
                                   &                    & \multicolumn{1}{c|}{VQ}       & 60.94 & 49.87  & 64.83 & 61.2      \\
                                   &                    & \multicolumn{1}{c|}{KBVQ-MoE} & 68.19 & 50.11  & 78.92 & 66.89     \\ \hline
\multirow{5}{*}{DeepSeekMoE-16B}   & 16                 & \multicolumn{1}{c|}{FP16}     & 44.60 & 31.49  & 20.16 & 26.83     \\ \cline{2-7} 
                                   & \multirow{4}{*}{2} & \multicolumn{1}{c|}{GPTQ}     & 28.48 & 18.82  & 11.26 & 14.87     \\
                                   &                    & \multicolumn{1}{c|}{MoeQuant} & 34.82 & 24.93  & 12.88 & 16.95     \\
                                   &                    & \multicolumn{1}{c|}{VQ}       & 30.78 & 24.49  & 13.55 & 17.89     \\
                                   &                    & \multicolumn{1}{c|}{KBVQ-MoE} & 41.39 & 28.97  & 17.32 & 22.62     \\ \hline
\multirow{5}{*}{Mixtral-7x8B}      & 16                 & \multicolumn{1}{c|}{FP16}     & 70.50 & 42.41  & 65.88 & 32.93     \\ \cline{2-7} 
                                   & \multirow{4}{*}{2} & \multicolumn{1}{c|}{GPTQ}     & 40.80 & 22.96  & 5.89  & 9.97      \\
                                   &                    & \multicolumn{1}{c|}{MoeQuant} & 49.39 & 27.11  & 20.67 & 14.74     \\
                                   &                    & \multicolumn{1}{c|}{VQ}       & 55.84 & 30.83  & 49.4  & 23.21     \\
                                   &                    & \multicolumn{1}{c|}{KBVQ-MoE} & 61.11 & 34.8   & 57.86 & 29.9      \\ \hline
\end{tabular}
}
\end{table}

\subsection{Limitations}
Despite the strong empirical performance of KBVQ-MoE on ultra–low-bit quantization of decoder-only MoE large language models, the method exhibits several limitations that highlight potential directions for future work.

(1) Dependence on empirically selected SVD truncation rank.
The truncated rank (k) in IDRE is currently chosen based on an empirical balance between reconstruction fidelity and storage overhead (e.g., $k = \frac{1}{128}$ of the full rank in our experiments). Although this choice works robustly across evaluated models, KBVQ-MoE does not yet include an adaptive mechanism to automatically determine the optimal rank for different MoE architectures, task regimes, or input statistics. This limits its ability to fully optimize the trade-off between redundancy removal and quantization efficiency across diverse settings.

(2) Limited validation beyond decoder-only architectures.
Our experiments focus primarily on decoder-only MoE models such as Qwen-MoE and Mixtral-8×7B. While this allows for a clean examination of the effects of IDRE and BCOS, the method has not yet been systematically evaluated on encoder–decoder MoE models or multimodal MoE architectures. These settings may require modified input-statistics estimation or revised forms of redundancy modeling. Extending KBVQ-MoE to bidirectional or cross-modal MoE structures is therefore an important area for future exploration.

(3) Lack of evaluation in extreme bit regimes.
KBVQ-MoE achieves near–FP16 accuracy at 2–3 bits, but has not been tested in more extreme quantization regimes such as 1-bit binary quantization or hybrid bit-widths (e.g., 1.5-bit). These regimes pose substantially greater information bottlenecks, and may require enhanced bias-correction mechanisms, more expressive codebook structures, or hybrid compression strategies to remain effective.

Together, these limitations delineate the boundaries of the current investigation and point toward promising extensions of KBVQ-MoE in future research.

%% file: tab/table5.tex
\begin{table}[!h]
\centering
\caption{Comparison experiment of the Qwen1.5-MoE-A2.7B model}
\label{tab5:qwen15-moe-a27b}
\resizebox{0.8\textwidth}{!}{
\begin{tabular}{c|c|ccccccccc}
\hline
Bit &
  Method &
  W2 &
  ARC-E &
  ARC-C &
  HE &
  LAMBADA-O &
  LAMBADA-S &
  PIQA &
  WI &
  AVG \\ \hline
16 &
  FP16 &
  7.22 &
  69.53 &
  44.03 &
  77.26 &
  71.28 &
  64.62 &
  80.47 &
  69.30 &
  68.07 \\ \hline
 &
  RTN &
  638509 &
  24.87 &
  27.90 &
  26.39 &
  0 &
  0 &
  51.47 &
  48.86 &
  25.64 \\
 &
  GPTQ &
  12.69 &
  47.14 &
  30.89 &
  60.77 &
  43.72 &
  34.81 &
  69.97 &
  56.20 &
  49.07 \\
 &
  MoeQuant &
  583542 &
  34.54 &
  35.09 &
  35.55 &
  12.03 &
  8.05 &
  59.08 &
  58.21 &
  34.64 \\
 &
  VQ &
  26.95 &
  58.16 &
  34.64 &
  63.35 &
  28.90 &
  22.92 &
  71.82 &
  55.09 &
  47.84 \\
\multirow{-5}{*}{2} &
  \cellcolor[HTML]{F2F3F5}KBVQ-MoE &
  \cellcolor[HTML]{F2F3F5}9.61 &
  \cellcolor[HTML]{F2F3F5}65.66 &
  \cellcolor[HTML]{F2F3F5}39.59 &
  \cellcolor[HTML]{F2F3F5}64.52 &
  \cellcolor[HTML]{F2F3F5}67.16 &
  \cellcolor[HTML]{F2F3F5}60.20 &
  \cellcolor[HTML]{F2F3F5}77.20 &
  \cellcolor[HTML]{F2F3F5}65.11 &
  \cellcolor[HTML]{F2F3F5}62.78 \\ \hline
 &
  RTN &
  116689 &
  25.84 &
  24.66 &
  27.52 &
  0.14 &
  0.02 &
  51.69 &
  49.88 &
  25.68 \\
 &
  GPTQ &
  7.58 &
  68.60 &
  43.34 &
  75.35 &
  68.68 &
  62.80 &
  79.22 &
  66.54 &
  66.36 \\
 &
  MoeQuant &
  7.87 &
  68.56 &
  44.21 &
  74.29 &
  70.54 &
  64.97 &
  79.51 &
  65.45 &
  66.79 \\
 &
  VQ &
  11.47 &
  60.86 &
  36.18 &
  60.56 &
  54.67 &
  41.30 &
  75.57 &
  62.43 &
  55.94 \\
\multirow{-5}{*}{3} &
  \cellcolor[HTML]{F2F3F5}KBVQ-MoE &
  \cellcolor[HTML]{F2F3F5}7.74 &
  \cellcolor[HTML]{F2F3F5}68.90 &
  \cellcolor[HTML]{F2F3F5}44.54 &
  \cellcolor[HTML]{F2F3F5}75.08 &
  \cellcolor[HTML]{F2F3F5}72.81 &
  \cellcolor[HTML]{F2F3F5}66.64 &
  \cellcolor[HTML]{F2F3F5}80.14 &
  \cellcolor[HTML]{F2F3F5}67.80 &
  \cellcolor[HTML]{F2F3F5}67.99 \\ \hline
\end{tabular}
}
\end{table}

%% file: tab/table6.tex
\begin{table}[!h]
\centering
\caption{Comparison experiment of the Qwen3-30B-A3B model}
\label{tab6:qwen3-30b-a3b}
\resizebox{0.8\textwidth}{!}{
\begin{tabular}{c|c|ccccccccc}
\hline
Bit & Method   & W2     & ARC-E & ARC-C & HE    & LAMBADA-O & LAMBADA-S & PIQA  & WI    & AVG   \\ \hline
16  & FP16     & 8.70   & 79.25 & 56.40 & 79.60 & 64.70     & 62.91     & 80.30 & 70.48 & 70.24 \\ \hline
    & RTN      & 765922 & 24.49 & 26.71 & 25.93 & 0         & 0         & 51.31 & 52.80 & 25.89 \\
    & GPTQ     & 438.42 & 34.49 & 27.85 & 33.89 & 14.62     & 18.11     & 52.23 & 50.23 & 33.06 \\
    & MoeQuant & 37465  & 29.81 & 26.59 & 25.38 & 7.93      & 11.78     & 50.74 & 50.41 & 28.94 \\
    & VQ       & 115.30 & 33.08 & 24.49 & 35.74 & 6.79      & 7.35      & 56.86 & 49.96 & 30.61 \\
\multirow{-5}{*}{2} &
  \cellcolor[HTML]{F2F3F5}KBVQ-MoE &
  \cellcolor[HTML]{F2F3F5}11.87 &
  \cellcolor[HTML]{F2F3F5}70.33 &
  \cellcolor[HTML]{F2F3F5}47.61 &
  \cellcolor[HTML]{F2F3F5}67.49 &
  \cellcolor[HTML]{F2F3F5}60.68 &
  \cellcolor[HTML]{F2F3F5}54.34 &
  \cellcolor[HTML]{F2F3F5}76.66 &
  \cellcolor[HTML]{F2F3F5}66.46 &
  \cellcolor[HTML]{F2F3F5}63.37 \\ \hline
    & RTN      & 68657  & 27.48 & 24.91 & 27.38 & 0.02      & 0.21      & 51.80 & 49.09 & 25.84 \\
    & GPTQ     & 11.32  & 75.92 & 50.29 & 71.32 & 59.83     & 53.93     & 69.12 & 64.19 & 63.51 \\
    & MoeQuant & 24.85  & 30.49 & 37.77 & 31.49 & 18.38     & 35.11     & 55.82 & 52.23 & 37.32 \\
    & VQ       & 18.72  & 57.83 & 40.87 & 63.23 & 40.44     & 32.51     & 71.82 & 57.70 & 52.06 \\
\multirow{-5}{*}{3} &
  \cellcolor[HTML]{F2F3F5}KBVQ-MoE &
  \cellcolor[HTML]{F2F3F5}9.26 &
  \cellcolor[HTML]{F2F3F5}77.27 &
  \cellcolor[HTML]{F2F3F5}53.24 &
  \cellcolor[HTML]{F2F3F5}75.53 &
  \cellcolor[HTML]{F2F3F5}66.37 &
  \cellcolor[HTML]{F2F3F5}62.35 &
  \cellcolor[HTML]{F2F3F5}78.89 &
  \cellcolor[HTML]{F2F3F5}70.01 &
  \cellcolor[HTML]{F2F3F5}69.09 \\ \hline
\end{tabular}
}
\end{table}

%% file: tab/table7.tex
\begin{table}[!h]
\centering
\caption{Comparison experiment of the Mixtral-8x7B model}
\label{tab7:mixtral-8x7b}
\resizebox{0.8\textwidth}{!}{
\begin{tabular}{c|c|ccccccccc}
\hline
Bit &
  Method &
  W2 &
  ARC-E &
  ARC-C &
  HE &
  LAMBADA-O &
  LAMBADA-S &
  PIQA &
  WI &
  AVG \\ \hline
16 &
  FP16 &
  3.88 &
  85.39 &
  66.38 &
  85.95 &
  77.28 &
  73.06 &
  85.20 &
  76.72 &
  78.57 \\ \hline
 &
  RTN &
  274952 &
  25.38 &
  28.41 &
  26.09 &
  0 &
  0 &
  50.76 &
  46.25 &
  25.27 \\
 &
  GPTQ &
  5.69 &
  72.35 &
  48.89 &
  76.95 &
  68.39 &
  61.44 &
  77.15 &
  67.72 &
  67.56 \\
 &
  MoeQuant &
  13.43 &
  49.85 &
  38.93 &
  40.12 &
  22.98 &
  18.94 &
  60.38 &
  49.91 &
  40.16 \\
 &
  VQ &
  5.99 &
  72.73 &
  44.62 &
  70.08 &
  50.46 &
  36.70 &
  74.42 &
  65.51 &
  59.22 \\
\multirow{-5}{*}{2} &
  \cellcolor[HTML]{F2F3F5}KBVQ-MoE &
  \cellcolor[HTML]{F2F3F5}5.39 &
  \cellcolor[HTML]{F2F3F5}81.52 &
  \cellcolor[HTML]{F2F3F5}58.19 &
  \cellcolor[HTML]{F2F3F5}80.62 &
  \cellcolor[HTML]{F2F3F5}80.57 &
  \cellcolor[HTML]{F2F3F5}72.48 &
  \cellcolor[HTML]{F2F3F5}82.59 &
  \cellcolor[HTML]{F2F3F5}73.88 &
  \cellcolor[HTML]{F2F3F5}75.69 \\ \hline
 &
  RTN &
  45136 &
  26.39 &
  28.50 &
  25.58 &
  0 &
  0 &
  51.58 &
  48.54 &
  25.80 \\
 &
  GPTQ &
  4.17 &
  84.01 &
  64.42 &
  85.12 &
  76.77 &
  71.76 &
  83.79 &
  76.16 &
  77.43 \\
 &
  MoeQuant &
  5.45 &
  80.23 &
  57.39 &
  80.91 &
  71.18 &
  64.95 &
  78.91 &
  71.91 &
  72.21 \\
 &
  VQ &
  5.52 &
  81.65 &
  58.45 &
  82.12 &
  72.97 &
  67.44 &
  81.88 &
  73.48 &
  73.98 \\
\multirow{-5}{*}{3} &
  \cellcolor[HTML]{F2F3F5}KBVQ-MoE &
  \cellcolor[HTML]{F2F3F5}4.07 &
  \cellcolor[HTML]{F2F3F5}84.89 &
  \cellcolor[HTML]{F2F3F5}63.05 &
  \cellcolor[HTML]{F2F3F5}84.99 &
  \cellcolor[HTML]{F2F3F5}80.50 &
  \cellcolor[HTML]{F2F3F5}75.53 &
  \cellcolor[HTML]{F2F3F5}83.90 &
  \cellcolor[HTML]{F2F3F5}75.61 &
  \cellcolor[HTML]{F2F3F5}78.35 \\ \hline
\end{tabular}
}
\end{table}

%% file: tab/table8.tex
\begin{table}[!h]
\centering
\caption{Comparison experiment of the DeepSeekV2-Lite model}
\label{tab8:deepseekv2-lite}
\resizebox{0.8\textwidth}{!}{
\begin{tabular}{c|c|ccccccccc}
\hline
Bit &
  Method &
  W2 &
  ARC-E &
  ARC-C &
  HE &
  LAMBADA-O &
  LAMBADA-S &
  PIQA &
  WI &
  AVG \\ \hline
16 &
  FP16 &
  5.92 &
  76.22 &
  48.98 &
  77.91 &
  72.33 &
  67.90 &
  80.20 &
  71.19 &
  70.68 \\ \hline
 &
  RTN &
  174653 &
  24.12 &
  26.45 &
  26.30 &
  0 &
  0 &
  49.89 &
  49.09 &
  25.12 \\
 &
  GPTQ &
  8.49 &
  63.47 &
  37.63 &
  65.45 &
  52.53 &
  48.55 &
  74.59 &
  64.09 &
  58.04 \\
 &
  MoeQuant &
  25893 &
  26.94 &
  25.40 &
  25.45 &
  0 &
  0 &
  50.14 &
  51.29 &
  25.59 \\
 &
  VQ &
  10.96 &
  53.99 &
  38.71 &
  50.92 &
  48.19 &
  40.11 &
  57.73 &
  59.33 &
  49.85 \\
\multirow{-5}{*}{2} &
  \cellcolor[HTML]{F2F3F5}KBVQ-MoE &
  \cellcolor[HTML]{F2F3F5}7.94 &
  \cellcolor[HTML]{F2F3F5}68.47 &
  \cellcolor[HTML]{F2F3F5}40.60 &
  \cellcolor[HTML]{F2F3F5}67.93 &
  \cellcolor[HTML]{F2F3F5}64.91 &
  \cellcolor[HTML]{F2F3F5}55.94 &
  \cellcolor[HTML]{F2F3F5}75.91 &
  \cellcolor[HTML]{F2F3F5}67.91 &
  \cellcolor[HTML]{F2F3F5}63.10 \\ \hline
 &
  RTN &
  97.75 &
  43.64 &
  27.05 &
  42.56 &
  2.68 &
  1.84 &
  66.00 &
  49.49 &
  33.32 \\
 &
  GPTQ &
  6.98 &
  74.04 &
  46.35 &
  76.44 &
  70.4 &
  65.65 &
  79.05 &
  70.27 &
  68.89 \\
 &
  MoeQuant &
  7.52 &
  70.25 &
  42.91 &
  71.19 &
  70.04 &
  65.32 &
  76.84 &
  67.83 &
  66.34 \\
 &
  VQ &
  7.94 &
  67.91 &
  43.22 &
  68.81 &
  67.90 &
  60.30 &
  67.82 &
  59.83 &
  62.26 \\
\multirow{-5}{*}{3} &
  \cellcolor[HTML]{F2F3F5}KBVQ-MoE &
  \cellcolor[HTML]{F2F3F5}6.95 &
  \cellcolor[HTML]{F2F3F5}73.74 &
  \cellcolor[HTML]{F2F3F5}45.73 &
  \cellcolor[HTML]{F2F3F5}74.36 &
  \cellcolor[HTML]{F2F3F5}72.62 &
  \cellcolor[HTML]{F2F3F5}66.33 &
  \cellcolor[HTML]{F2F3F5}79.05 &
  \cellcolor[HTML]{F2F3F5}69.30 &
  \cellcolor[HTML]{F2F3F5}68.73 \\ \hline
\end{tabular}
}
\end{table}

%% file: algo/algo.tex
\subsection{KBVQ-MoE Algorithm}\label{appendix:algo}
\algrenewcommand{\algorithmicrequire}{\textbf{Input:}}
\algrenewcommand{\algorithmicensure}{\textbf{Output:}}

\newcommand{\Module}[2]{%
  \vspace{0.8em}%
  \Statex \colorbox{#1!10}{\parbox{\linewidth-2\fboxsep}{\textbf{#2}}}%
}

\begin{algorithm}[H]
\caption{KBVQ-MoE: KLT-guided SVD with Bias-Corrected VQ for MoE Models}
\label{alg:kbvq_moe}
\begin{algorithmic}[1]
\Require Expert weight matrices $\{W^{(i)}\}_{i=1}^n$, input activations $X$, codebook size $K$, sub-vector length $d$
\Ensure Quantized weights $\{\hat{W}^{(i)}\}_{i=1}^n$ with bias correction parameters $(s,b)$

\Module{blue}{\textbf{Pre-Process: KLT-guided SVD (Redundancy Removal)}}
\State Compute input covariance: $C_X=\tfrac{1}{B-1}X^\top X \in\mathbb{R}^{ic\times ic}$
\State Eigen-decompose: $C_X=U_{\mathrm{KLT}}^\top\Lambda_{\mathrm{KLT}}U_{\mathrm{KLT}}$
\State Input-coherent basis: $U_X=U_{\mathrm{KLT}}\Lambda_{\mathrm{KLT}}^{1/2}$;
\For{$i=1$ \textbf{to} $n$}
  \State Project expert weights (right projection): $\widetilde W^{(i)}=W^{(i)}U_X$
\EndFor
\State Stack all experts (row-wise): $\bar W=[\widetilde W^{(1)};\cdots;\widetilde W^{(n)}]\in\mathbb{R}^{(n*oc)\times ic}$
\State Product SVD: $\bar W=(U\Sigma V^\top)^T$ \Comment{$V\in\mathbb{R}^{(oc)\times n*oc}$, $U\in\mathbb{R}^{ic\times ic}$}
\State Select top-$k$: $U_k=U_{[:,1:k]}$, $V_k=V_{[:,1:k]}$, $\Sigma_k=\Sigma_{1:k,1:k}$
\State Partition $V_k$ by experts: $V_k=[\Sigma_k V_k^{(1)};\ldots;\Sigma_k V_k^{(n)}]$, each $V_k^{(i)}\in\mathbb{R}^{oc\times k}$
\For{$i=1$ \textbf{to} $n$}
  \State Define shared output loader: $U_{\text{share}}\gets  U_X^{-1} U_k$ \Comment{$ic\times k$}
  \State Shared part in projected basis: $\widehat W_{\text{share}}^{(i)}=(U_{\text{share}} (V^{(i)}_k)^T)^T$ \Comment{$oc\times ic$}
  \State Map back to original input space: $W_{\text{share}}^{(i)}=\widehat W_{\text{share}}^{(i)}\,$
  \State Special (residual) part: $W_{\text{quant}}^{(i)}=W^{(i)}-W_{\text{share}}^{(i)}$
\EndFor

\Module{red}{\textbf{Quantization: Vector Quantization of Special Part}}
\For{$i = 1$ to $n$}
    \State Split $W^{(i)}_{\text{quant}}$ into sub-vectors $\{z\}$
    \State Initialize codebook via K-means++
    \State Train VQ codebook $C = \{c_1,\dots,c_K\}$ by k-means
    \For{each sub-vector $z$}
        \State Assign index: $q = \arg\min_j \|z - c_j\|^2$
        \State Replace: $z_q = c_q$
    \EndFor
    \State Form quantized special part: $W^{(i)}_{\text{quant,VQ}}$
\EndFor

\Module{green}{\textbf{Post-Process: Bias Correction}}
\For{$i = 1$ to $n$}
    \State Define quantized weights: $\hat{W}^{(i)} = W^{(i)}_{\text{share}} + W^{(i)}_{\text{quant,VQ}}$
    \State Estimate per-channel statistics from calibration data:
    \Statex \hspace{1.5em} $\mu_{y}, \sigma_{y}$ for original outputs $y = W^{(i)}x$
    \Statex \hspace{1.5em} $\mu_{\hat{y}}, \sigma_{\hat{y}}$ for quantized outputs $\hat{y} = \hat{W}^{(i)}x$
    \State Compute correction parameters:
    \Statex \hspace{1.5em} $s_j = \tfrac{\sigma_{y_j}}{\sigma_{\hat{y}_j}} - 1$, \quad $b_j = \mu_{y_j} - (1+s_j)\mu_{\hat{y}_j}$
    \State Corrected output: $y_{\text{corr}} = (1+s) \odot (\hat{W}^{(i)}x) + b$
\EndFor

\State \Return   \textbf{$U_{\text{share}}$}, \textbf{$V_k$}, $C$, $(s,b)$
\end{algorithmic}
\end{algorithm}

%% file: iclr2026/iclr2026_conference.bib
@article{gpt4,
  title        = {GPT-4 Technical Report},
  author       = {{OpenAI}},
  journal      = {CoRR},
  volume       = {abs/2303.08774},
  year         = {2023},
  doi          = {10.48550/arXiv.2303.08774},
  url          = {https://doi.org/10.48550/arXiv.2303.08774},
}

@article{deepseekv2,
  title        = {DeepSeek-V2: A Strong, Economical, and Efficient Mixture-of-Experts Language Model},
  author       = {DeepSeek-AI and Aixin Liu and Bei Feng and Bin Wang and Bingxuan Wang and Bo Liu and Chenggang Zhao and Chengqi Deng and Chong Ruan and Damai Dai and Daya Guo and Dejian Yang and Deli Chen and Dongjie Ji and Erhang Li and Fangyun Lin and Fuli Luo and Guangbo Hao and Guanting Chen and Guowei Li and H. Zhang and Hanwei Xu and Hao Yang and Haowei Zhang and Honghui Ding and Huajian Xin and Huazuo Gao and Hui Li and Hui Qu and J. L. Cai and Jian Liang and Jianzhong Guo and Jiaqi Ni and Jiashi Li and Jin Chen and Jingyang Yuan and Junjie Qiu and Junxiao Song and Kai Dong and Kaige Gao and Kang Guan and Lean Wang and Lecong Zhang and Lei Xu and Leyi Xia and Liang Zhao and Liyue Zhang and Meng Li and Miaojun Wang and Mingchuan Zhang and Minghua Zhang and Minghui Tang and Mingming Li and Ning Tian and Panpan Huang and Peiyi Wang and Peng Zhang and Qihao Zhu and Qinyu Chen and Qiushi Du and R. J. Chen and R. L. Jin and Ruiqi Ge and Ruizhe Pan and Runxin Xu and Ruyi Chen and S. S. Li and Shanghao Lu and Shangyan Zhou and Shanhuang Chen and Shaoqing Wu and Shengfeng Ye and Shirong Ma and Shiyu Wang and Shuang Zhou and Shuiping Yu and Shunfeng Zhou and Size Zheng and T. Wang and Tian Pei and Tian Yuan and Tianyu Sun and W. L. Xiao and Wangding Zeng and Wei An and Wen Liu and Wenfeng Liang and Wenjun Gao and Wentao Zhang and X. Q. Li and Xiangyue Jin and Xianzu Wang and Xiao Bi and Xiaodong Liu and Xiaohan Wang and Xiaojin Shen and Xiaokang Chen and Xiaosha Chen and Xiaotao Nie and Xiaowen Sun and Xiaoxiang Wang and Xin Liu and Xin Xie and Xingkai Yu and Xinnan Song and Xinyi Zhou and Xinyu Yang and Xuan Lu and Xuecheng Su and Y. Wu and Y. K. Li and Y. X. Wei and Y. X. Zhu and Yanhong Xu and Yanping Huang and Yao Li and Yao Zhao and Yaofeng Sun and Yaohui Li and Yaohui Wang and Yi Zheng and Yichao Zhang and Yiliang Xiong and Yilong Zhao and Ying He and Ying Tang and Yishi Piao and Yixin Dong and Yixuan Tan and Yiyuan Liu and Yongji Wang and Yongqiang Guo and Yuchen Zhu and Yuduan Wang and Yuheng Zou and Yukun Zha and Yunxian Ma and Yuting Yan and Yuxiang You and Yuxuan Liu and Z. Z. Ren and Zehui Ren and Zhangli Sha and Zhe Fu and Zhen Huang and Zhen Zhang and Zhenda Xie and Zhewen Hao and Zhihong Shao and Zhiniu Wen and Zhipeng Xu and Zhongyu Zhang and Zhuoshu Li and Zihan Wang and Zihui Gu and Zilin Li and Ziwei Xie},
  journal      = {CoRR},
  volume       = {abs/2405.04434},
  year         = {2024},
  url          = {https://arxiv.org/abs/2405.04434},
}

@article{mixtral,
  title        = {Mixtral of Experts},
  author       = {Albert Q. Jiang and Alexandre Sablayrolles and Antoine Roux and Arthur Mensch and Blanche Savary and Chris Bamford and Devendra Singh Chaplot and Diego de las Casas and Emma Bou Hanna and Florian Bressand and Gianna Lengyel and Guillaume Bour and Guillaume Lample and Lélio Renard Lavaud and Lucile Saulnier and Marie-Anne Lachaux and Pierre Stock and Sandeep Subramanian and Sophia Yang and Szymon Antoniak and Teven Le Scao and Théophile Gervet and Thibaut Lavril and Thomas Wang and Timothée Lacroix and William El Sayed},
  journal      = {CoRR},
  volume       = {abs/2401.04088},
  year         = {2024},
  url          = {https://arxiv.org/abs/2401.04088},
}

@article{qwen3,
  title        = {Qwen3 Technical Report},
  author       = {Yang, An and Li, Anfeng and Yang, Baosong and Zhang, Beichen and Hui, Binyuan and Zheng, Bo and Yu, Bowen and Gao, Chang and Huang, Chengen and Lv, Chenxu and Zheng, Chujie and Liu, Dayiheng and Zhou, Fan and Huang, Fei and Hu, Feng and Ge, Hao and Wei, Haoran and Lin, Huan and Tang, Jialong and Yang, Jian and Tu, Jianhong and Zhang, Jianwei and Yang, Jianxin and Yang, Jiaxi and Zhou, Jing and Zhou, Jingren and Lin, Junyang and Dang, Kai and Bao, Keqin and Yang, Kexin and Yu, Le and Deng, Lianghao and Li, Mei and Xue, Mingfeng and Li, Mingze and Zhang, Pei and Wang, Peng and Zhu, Qin and Men, Rui and Gao, Ruize and Liu, Shixuan and Luo, Shuang and Li, Tianhao and Tang, Tianyi and Yin, Wenbiao and Ren, Xingzhang and Wang, Xinyu and Zhang, Xinyu and Ren, Xuancheng and Fan, Yang and Su, Yang and Zhang, Yichang and Zhang, Yinger and Wan, Yu and Liu, Yuqiong and Wang, Zekun and Cui, Zeyu and Zhang, Zhenru and Zhou, Zhipeng and Qiu, Zihan and others},
  journal      = {CoRR},
  volume       = {abs/2505.09388},
  year         = {2025},
  url          = {https://arxiv.org/abs/2505.09388},
}

@article{kimivl,
  title        = {Kimi-VL Technical Report},
  author       = {{Kimi Team}},
  journal      = {CoRR},
  volume       = {abs/2504.07491},
  year         = {2025},
  url          = {https://arxiv.org/abs/2504.07491},
}

@article{gptq,
  title={Gptq: Accurate post-training quantization for generative pre-trained transformers},
  author={Frantar, Elias and Ashkboos, Saleh and Hoefler, Torsten and Alistarh, Dan},
  journal={arXiv preprint arXiv:2210.17323},
  year={2022}
}

@article{ostquant,
  title={Ostquant: Refining large language model quantization with orthogonal and scaling transformations for better distribution fitting},
  author={Hu, Xing and Cheng, Yuan and Yang, Dawei and Xu, Zukang and Yuan, Zhihang and Yu, Jiangyong and Xu, Chen and Jiang, Zhe and Zhou, Sifan},
  journal={arXiv preprint arXiv:2501.13987},
  year={2025}
}

@article{pcdvq,
  title={PCDVQ: Enhancing Vector Quantization for Large Language Models via Polar Coordinate Decoupling},
  author={Yue, Yuxuan and Xu, Zukang and Yuan, Zhihang and Yang, Dawei and Wu, Jianlong and Nie, Liqiang},
  journal={arXiv preprint arXiv:2506.05432},
  year={2025}
}

@article{vptq,
  title={Vptq: Extreme low-bit vector post-training quantization for large language models},
  author={Liu, Yifei and Wen, Jicheng and Wang, Yang and Ye, Shengyu and Zhang, Li Lyna and Cao, Ting and Li, Cheng and Yang, Mao},
  journal={arXiv preprint arXiv:2409.17066},
  year={2024}
}

@article{qtip,
  title={Qtip: Quantization with trellises and incoherence processing},
  author={Tseng, Albert and Sun, Qingyao and Hou, David and De Sa, Christopher M},
  journal={Advances in Neural Information Processing Systems},
  volume={37},
  pages={59597--59620},
  year={2024}
}

@article{moeadaptation,
  title={Mixture-of-Experts for Open Set Domain Adaptation: A Dual-Space Detection Approach},
  author={Du, Zhenbang and An, Jiayu and Tu, Yunlu and Hong, Jiahao and Wu, Dongrui},
  journal={IEEE Transactions on Artificial Intelligence},
  year={2025},
  publisher={IEEE}
}

@article{sankar,
  title={Mixture of Experts Models in Deep Learning and Their Techniques Applications and Challenges},
  author={Sankar, Eustache and Dimitri, Vasily},
  journal={Authorea Preprints}, 
  year={2025},
  publisher={Authorea}
}

@article{load,
  title={Load Balancing Mixture of Experts with Similarity Preserving Routers},
  author={Omi, Nabil and Sen, Siddhartha and Farhadi, Ali},
  journal={arXiv preprint arXiv:2506.14038},
  year={2025}
}

@article{d2moe,
  title={Delta decompression for moe-based llms compression},
  author={Gu, Hao and Li, Wei and Li, Lujun and Zhu, Qiyuan and Lee, Mark and Sun, Shengjie and Xue, Wei and Guo, Yike},
  journal={arXiv preprint arXiv:2502.17298},
  year={2025}
}

@article{submoe,
  title={Sub-MoE: Efficient Mixture-of-Expert LLMs Compression via Subspace Expert Merging},
  author={Li, Lujun and Qiyuan, Zhu and Wang, Jiacheng and Li, Wei and Gu, Hao and Han, Sirui and Guo, Yike},
  journal={arXiv preprint arXiv:2506.23266},
  year={2025}
}

@article{quarot,
  title={Quarot: Outlier-free 4-bit inference in rotated llms},
  author={Ashkboos, Saleh and Mohtashami, Amirkeivan and Croci, Maximilian L and Li, Bo and Cameron, Pashmina and Jaggi, Martin and Alistarh, Dan and Hoefler, Torsten and Hensman, James},
  journal={Advances in Neural Information Processing Systems},
  volume={37},
  pages={100213--100240},
  year={2024}
}

@article{mambaquant,
  title={Mambaquant: Quantizing the mamba family with variance aligned rotation methods},
  author={Xu, Zukang and Yue, Yuxuan and Hu, Xing and Yuan, Zhihang and Jiang, Zixu and Chen, Zhixuan and Yu, Jiangyong and Xu, Chen and Zhou, Sifan and Yang, Dawei},
  journal={arXiv preprint arXiv:2501.13484},
  year={2025}
}

@article{gptaq,
  title={GPTAQ: Efficient Finetuning-Free Quantization for Asymmetric Calibration},
  author={Li, Yuhang and Yin, Ruokai and Lee, Donghyun and Xiao, Shiting and Panda, Priyadarshini},
  journal={arXiv preprint arXiv:2504.02692},
  year={2025}
}

@article{illm,
  title={I-llm: Efficient integer-only inference for fully-quantized low-bit large language models},
  author={Hu, Xing and Cheng, Yuan and Yang, Dawei and Yuan, Zhihang and Yu, Jiangyong and Xu, Chen and Zhou, Sifan},
  journal={arXiv preprint arXiv:2405.17849},
  year={2024}
}

@article{moequant,
  title={MoEQuant: Enhancing Quantization for Mixture-of-Experts Large Language Models via Expert-Balanced Sampling and Affinity Guidance},
  author={Hu, Xing and Chen, Zhixuan and Yang, Dawei and Xu, Zukang and Xu, Chen and Yuan, Zhihang and Zhou, Sifan and Yu, Jiangyong},
  journal={arXiv preprint arXiv:2505.03804},
  year={2025}
}

@article{vector_quantize,
  title={Asymptotically optimal block quantization},
  author={Gersho, Allen},
  journal={IEEE Transactions on information theory},
  volume={25},
  number={4},
  pages={373--380},
  year={1979},
  publisher={IEEE}
}

@article{gptvq,
  title={Gptvq: The blessing of dimensionality for llm quantization},
  author={Van Baalen, Mart and Kuzmin, Andrey and Koryakovskiy, Ivan and Nagel, Markus and Couperus, Peter and Bastoul, Cedric and Mahurin, Eric and Blankevoort, Tijmen and Whatmough, Paul},
  journal={arXiv preprint arXiv:2402.15319},
  year={2024}
}

@article{quip_sharp,
  title={Quip\#: Even better llm quantization with hadamard incoherence and lattice codebooks},
  author={Tseng, Albert and Chee, Jerry and Sun, Qingyao and Kuleshov, Volodymyr and De Sa, Christopher},
  journal={arXiv preprint arXiv:2402.04396},
  year={2024}
}

@article{aqlm,
  title={Extreme compression of large language models via additive quantization},
  author={Egiazarian, Vage and Panferov, Andrei and Kuznedelev, Denis and Frantar, Elias and Babenko, Artem and Alistarh, Dan},
  journal={arXiv preprint arXiv:2401.06118},
  year={2024}
}

@article{eacmoe,
  title={EAC-MoE: Expert-Selection Aware Compressor for Mixture-of-Experts Large Language Models},
  author={Chen, Yuanteng and Shao, Yuantian and Wang, Peisong and Cheng, Jian},
  journal={arXiv preprint arXiv:2508.01625},
  year={2025}
}

@misc{qwen1_5,
  title        = {Qwen1.5-MoE: Matching 7B Model Performance with 1/3 Activated Parameters},
  author       = {{Qwen Team}},
  howpublished = {\url{https://qwenlm.github.io/blog/qwen-moe/}},
  year         = {2024},
}

@article{wikitext2,
  title={Pointer sentinel mixture models},
  author={Merity, Stephen and Xiong, Caiming and Bradbury, James and Socher, Richard},
  journal={arXiv preprint arXiv:1609.07843},
  year={2016}
}

@article{arc-c,
  title={Think you have solved question answering? try arc, the ai2 reasoning challenge},
  author={Clark, Peter and Cowhey, Isaac and Etzioni, Oren and Khot, Tushar and Sabharwal, Ashish and Schoenick, Carissa and Tafjord, Oyvind},
  journal={arXiv preprint arXiv:1803.05457},
  year={2018}
}

@article{arc-e,
  title={A systematic classification of knowledge, reasoning, and context within the ARC dataset},
  author={Boratko, Michael and Padigela, Harshit and Mikkilineni, Divyendra and Yuvraj, Pritish and Das, Rajarshi and McCallum, Andrew and Chang, Maria and Fokoue-Nkoutche, Achille and Kapanipathi, Pavan and Mattei, Nicholas and others},
  journal={arXiv preprint arXiv:1806.00358},
  year={2018}
}

@article{hellaswag,
  title={Hellaswag: Can a machine really finish your sentence?},
  author={Zellers, Rowan and Holtzman, Ari and Bisk, Yonatan and Farhadi, Ali and Choi, Yejin},
  journal={arXiv preprint arXiv:1905.07830},
  year={2019}
}

@misc{lambda,
      title={The LAMBADA dataset: Word prediction requiring a broad discourse context}, 
      author={Denis Paperno and Germán Kruszewski and Angeliki Lazaridou and Quan Ngoc Pham and Raffaella Bernardi and Sandro Pezzelle and Marco Baroni and Gemma Boleda and Raquel Fernández},
      year={2016},
      eprint={1606.06031},
      archivePrefix={arXiv},
      primaryClass={cs.CL},
      url={https://arxiv.org/abs/1606.06031}, 
}

@inproceedings{piqa,
  title={Piqa: Reasoning about physical commonsense in natural language},
  author={Bisk, Yonatan and Zellers, Rowan and Gao, Jianfeng and Choi, Yejin},
  booktitle={Proceedings of the AAAI Conference on Artificial Intelligence},
  volume={34},
  pages={7432--7439},
  year={2020}
}

@article{winogrande,
  title={Winogrande: An adversarial winograd schema challenge at scale},
  author={Sakaguchi, Keisuke and Bras, Ronan Le and Bhagavatula, Chandra and Choi, Yejin},
  journal={Communications of the ACM},
  volume={64},
  number={9},
  pages={99--106},
  year={2021},
  publisher={ACM New York, NY, USA}
}

@misc{RedPajama,
  title = {RedPajama-Data-1T: An Open Dataset for Training Large Language Models},
  author = {{Together Computer}},
  year = {2023},
  howpublished = {\url{https://huggingface.co/datasets/togethercomputer/RedPajama-Data-1T}},
  note = {Available on Hugging Face Datasets}
}

@techreport{lm-eval,
  author       = {Leo Gao and Jonathan Tow and Stella Biderman and Sid Black and Anthony DiPofi and Charles Foster and Laurence Golding and Jeffrey Hsu and Kyle McDonell and Niklas Muennighoff},
  title        = {A Framework for Few-Shot Language Model Evaluation},
  year         = {2021},
  month        = sep,
  version      = {v0.0.1},
  note         = {},
  institution  = {GitHub},
  howpublished = {Available online at \url{https://github.com/EleutherAI/lm-evaluation-harness}}
}

@misc{qwen3-next,
  title   = {Qwen3-Next: Next-Generation Foundation Model Architecture},
  author  = {Alibaba Cloud, Tongyi Qianwen Team},
  howpublished = {\url{https://modelscope.cn/models/Qwen/Qwen3-Next-80B-A3B-Instruct}},
  year    = {2025},
}

@article{6797059,
  author={Jacobs, Robert A. and Jordan, Michael I. and Nowlan, Steven J. and Hinton, Geoffrey E.},
  journal={Neural Computation}, 
  title={Adaptive Mixtures of Local Experts}, 
  year={1991},
  volume={3},
  number={1},
  pages={79-87},
  doi={10.1162/neco.1991.3.1.79},
}

@article{6796382,
  author={Jordan, Michael I. and Jacobs, Robert A.},
  journal={Neural Computation}, 
  title={Hierarchical Mixtures of Experts and the EM Algorithm}, 
  year={1994},
  volume={6},
  number={2},
  pages={181-214},
  doi={10.1162/neco.1994.6.2.181}
}

@article{flatquant,
  title={Flatquant: Flatness matters for llm quantization},
  author={Sun, Yuxuan and Liu, Ruikang and Bai, Haoli and Bao, Han and Zhao, Kang and Li, Yuening and Hu, Jiaxin and Yu, Xianzhi and Hou, Lu and Yuan, Chun and others},
  journal={arXiv preprint arXiv:2410.09426},
  year={2024}
}

@article{mmlu,
  title={Mmlu-pro: A more robust and challenging multi-task language understanding benchmark},
  author={Wang, Yubo and Ma, Xueguang and Zhang, Ge and Ni, Yuansheng and Chandra, Abhranil and Guo, Shiguang and Ren, Weiming and Arulraj, Aaran and He, Xuan and Jiang, Ziyan and others},
  journal={Advances in Neural Information Processing Systems},
  volume={37},
  pages={95266--95290},
  year={2024}
}

@inproceedings{mathqa,
  title={Mathqa: Towards interpretable math word problem solving with operation-based formalisms},
  author={Amini, Aida and Gabriel, Saadia and Lin, Shanchuan and Koncel-Kedziorski, Rik and Choi, Yejin and Hajishirzi, Hannaneh},
  booktitle={Proceedings of the 2019 conference of the North American chapter of the association for computational linguistics: Human language technologies, volume 1 (long and short papers)},
  pages={2357--2367},
  year={2019}
}

@misc{gsm8k,
  title={Training Verifiers to Solve Math Word Problems},
  author={Karl Cobbe and Vineet Kosaraju and Mohammad Bavarian and Jacob Hilton and Reiichiro Nakano and Christopher Hesse and John Schulman},
  year={2021},
  eprint={2110.14168},
  archivePrefix={arXiv},
  primaryClass={cs.LG}
}

@misc{humaneval,
  title={Evaluating Large Language Models Trained on Code},
  author={Mark Chen and Jerry Tworek and Heewoo Jun and Qiming Yuan and Henrique Ponde de Oliveira Pinto and Jared Kaplan and Harri Edwards and Yuri Burda and Nicholas Joseph and Greg Brockman and Alex Ray and Raul Puri and Gretchen Krueger and Michael Petrov and Heidy Khlaaf and Girish Sastry and Pamela Mishkin and Brooke Chan and Scott Gray and Nick Ryder and Mikhail Pavlov and Alethea Power and Lukasz Kaiser and Mohammad Bavarian and Clemens Winter and Philippe Tillet and Felipe Petroski Such and Dave Cummings and Matthias Plappert and Fotios Chantzis and Elizabeth Barnes and Ariel Herbert-Voss and William H. Guss and Andrew N. Nichol and Alex Paino and Nikolas Tezak and Jie Tang and Igor Babuschkin and Sandeep Balaji and Sharan Jain and William Saunders and Christopher Hesse and John Schulman},
  year={2021},
  eprint={2107.03374},
  archivePrefix={arXiv},
  primaryClass={cs.CL}
}

@article{specquant,
  title={SpecQuant: Spectral Decomposition and Adaptive Truncation for Ultra-Low-Bit LLMs Quantization},
  author={Zhao, Zhixiong and Liu, Fangxin and Wang, Junjie and Guan, Chenyang and Wang, Zongwu and Jiang, Li and Guan, Haibing},
  journal={arXiv preprint arXiv:2511.11663},
  year={2025}
}
